%% file: main.tex
\documentclass[10pt,logo,copyright]{giga-report}
\input{packages}

\input{common.tex}

\definecolor{darkred}{rgb}{0.7, 0.0, 0.0}

\usepackage{makecell}
\usepackage{tabularx}
\usepackage{graphicx}
\usepackage{booktabs}
\usepackage{multirow}
\usepackage{listings}
\usepackage{wrapfig}
\usepackage{array}
\usepackage{enumitem}
\usepackage[table]{xcolor}
\definecolor{mygray}{gray}{0.9}

\usepackage{amsmath}
\usepackage{pifont}

\usepackage[nameinlink]{cleveref}
\crefname{equation}{Eq.}{Eqs.}
\crefname{figure}{Fig.}{Figs.}
\crefname{section}{Sec.}{Secs.}
\crefname{appendix}{App.}{Apps.}
\crefname{table}{Tab.}{Tabs.}
\crefname{algorithm}{Algo.}{Algos.}
\crefname{thm}{Thm.}{Thms.}
\Crefname{thm}{Thm.}{Thms.}
\crefname{prop}{Prop.}{Props.}

\newcommand{\crefnames}[3]{%
  \@for\next:=#1\do{%
    \expandafter\crefname\expandafter{\next}{#2}{#3}%
  }%
}

\title{GigaWorld-1: A Roadmap to Build World Models for Robot Policy Evaluation}

\author{
\vspace{-0.1in}
\centerline{GigaAI}
\centerline{Tsinghua University}
\centerline{{Project Page: \href{https://open-gigaai.github.io/giga-world-1/}{https://open-gigaai.github.io/giga-world-1/}}}
\footnotesize
\textbf{Alphabetical Order}:
\normalfont
  Angyuan Ma,
  Boyuan Wang,
  Bohan Li,
  Chaojun Ni,
  Guo Li,
  Guan Huang,
  Guosheng Zhao,
  Hao Li,
  Hengtao Li,
  Jingyu Liu,
  Jiwen Lu,
  Qiuping Deng,
  Tingdong Yu,
  Xuancheng Xu,
  Xinyu Zhou,
  Xiuwei Xu,
  Xinze Chen,
  Xiaofeng Wang,\newline
  Xiaoyu Tian,
  Yang Wang,
  Yifan Chang,
  Yukun Zhou,
  Yun Ye,
  Zhenyu Wu,
  Zhanqian Wu,
  Zheng Zhu
\vspace{-1.5em}
}
\usepackage[numbers,sort&compress]{natbib}
\setcitestyle{numbers}
\begin{document}
\maketitle

\begin{figure*}[htbp]
\centering
\captionsetup{type=figure, justification=justified, singlelinecheck=false}
\includegraphics[width=0.95\textwidth]{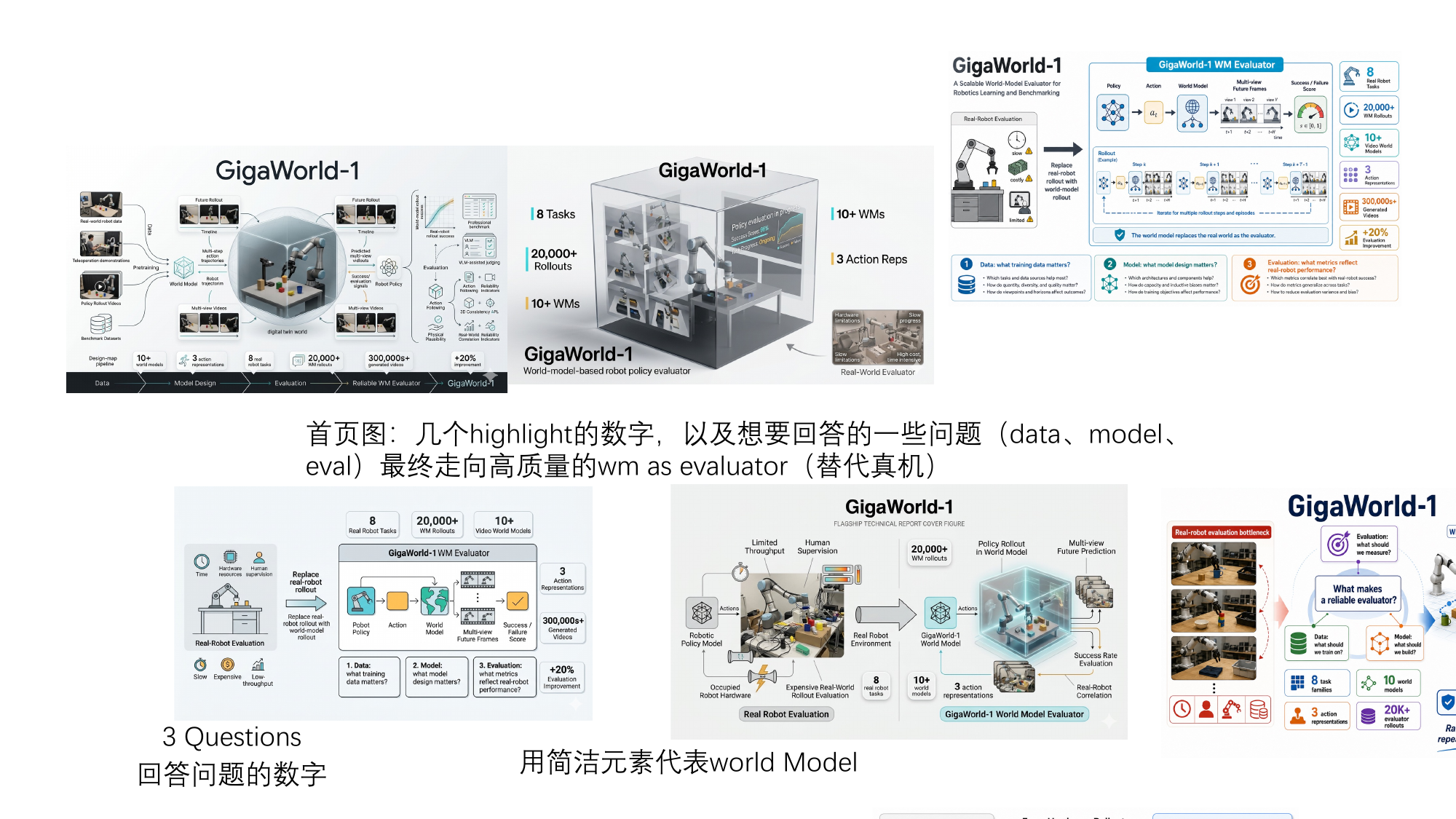}
\caption{This paper analyzes 324,000 world-model-simulated rollouts, 7 video world models, 4 action representation paradigms. Complemented by more than 12,000 hours of training data and guided by the roadmap we establish for building world models tailored to robot policy evaluation, we introduce \textit{GigaWorld-1}.}
\label{fig:teaser}
\end{figure*}

\vspace{-1.5em}

\begin{abstract}
\input{sections/abstract}
\end{abstract}
\abscontent

\input{sections/introduction}

\input{sections/related_work}

\input{sections/preliminaries}

\input{sections/wmbench}

\input{sections/what_matters}
\input{sections/design_map}
\input{sections/discussion}

\clearpage

\bibliographystyle{plainnat}
\bibliography{main}

\end{document}

%% file: packages.tex
\usepackage{iftex}
\ifPDFTeX
    \usepackage[utf8]{inputenc} %
    \usepackage[T1]{fontenc}    %
\fi

\usepackage{parskip}        %
\usepackage{url}            %
\usepackage{booktabs}       %
\usepackage{amsfonts}       %
\usepackage{nicefrac}       %
\usepackage{microtype}      %
\usepackage[dvipsnames]{xcolor} %
\usepackage{graphicx}
\usepackage{animate}        %
\usepackage{subcaption}
\usepackage{tabularx}
\usepackage{makecell}
\usepackage{adjustbox}
\usepackage{setspace}
\usepackage{todonotes}
\usepackage{colortbl}
\usepackage{wrapfig}
\usepackage{svg} 

\definecolor{rankbest}{RGB}{144,238,144}
\definecolor{rankmid}{RGB}{255,255,180}
\definecolor{rankworst}{RGB}{248,150,150}
\definecolor{deltadarkpos}{RGB}{50,180,50}
\definecolor{deltalightpos}{RGB}{180,235,180}
\definecolor{deltayellow}{RGB}{255,255,200}
\definecolor{deltalightneg}{RGB}{250,180,180}
\definecolor{deltadarkneg}{RGB}{220,80,80}

\newcolumntype{M}[1]{>{\centering\arraybackslash}m{#1}}
\usepackage{float}
\usepackage{tikz}
\usetikzlibrary{positioning,shapes,arrows,arrows.meta,fit,tikzmark}
\usepackage{amsmath,amsfonts,bm, bbm,leftindex}
\usepackage{multirow}
\usepackage{comment}
\usepackage{lipsum}
\usepackage[para]{threeparttable}

%% file: common.tex
\def\eqref#1{equation~\ref{#1}}

\def\1{\bm{1}}

\DeclareMathAlphabet{\mathsfit}{\encodingdefault}{\sfdefault}{m}{sl}
\SetMathAlphabet{\mathsfit}{bold}{\encodingdefault}{\sfdefault}{bx}{n}

\makeatletter
\let\save@mathaccent\mathaccent
\newcommand*\if@single[3]{%
  \setbox0\hbox{${\mathaccent"0362{#1}}^H$}%
  \setbox2\hbox{${\mathaccent"0362{\kern0pt#1}}^H$}%
  \ifdim\ht0=\ht2 #3\else #2\fi
  }
\newcommand*\rel@kern[1]{\kern#1\dimexpr\macc@kerna}
\newcommand*\widebar[1]{\@ifnextchar^{{\wide@bar{#1}{0}}}{\wide@bar{#1}{1}}}
\newcommand*\wide@bar[2]{\if@single{#1}{\wide@bar@{#1}{#2}{1}}{\wide@bar@{#1}{#2}{2}}}
\newcommand*\wide@bar@[3]{%
  \begingroup
  \def\mathaccent##1##2{%
    \let\mathaccent\save@mathaccent
    \if#32 \let\macc@nucleus\first@char \fi
    \setbox\z@\hbox{$\macc@style{\macc@nucleus}_{}$}%
    \setbox\tw@\hbox{$\macc@style{\macc@nucleus}{}_{}$}%
    \dimen@\wd\tw@
    \advance\dimen@-\wd\z@
    \divide\dimen@ 3
    \@tempdima\wd\tw@
    \advance\@tempdima-\scriptspace
    \divide\@tempdima 10
    \advance\dimen@-\@tempdima
    \ifdim\dimen@>\z@ \dimen@0pt\fi
    \rel@kern{0.6}\kern-\dimen@
    \if#31
      \overline{\rel@kern{-0.6}\kern\dimen@\macc@nucleus\rel@kern{0.4}\kern\dimen@}%
      \advance\dimen@0.4\dimexpr\macc@kerna
      \let\final@kern#2%
      \ifdim\dimen@<\z@ \let\final@kern1\fi
      \if\final@kern1 \kern-\dimen@\fi
    \else
      \overline{\rel@kern{-0.6}\kern\dimen@#1}%
    \fi
  }%
  \macc@depth\@ne
  \let\math@bgroup\@empty \let\math@egroup\macc@set@skewchar
  \mathsurround\z@ \frozen@everymath{\mathgroup\macc@group\relax}%
  \macc@set@skewchar\relax
  \let\mathaccentV\macc@nested@a
  \if#31
    \macc@nested@a\relax111{#1}%
  \else
    \def\gobble@till@marker##1\endmarker{}%
    \futurelet\first@char\gobble@till@marker#1\endmarker
    \ifcat\noexpand\first@char A\else
      \def\first@char{}%
    \fi
    \macc@nested@a\relax111{\first@char}%
  \fi
  \endgroup
}
\makeatother

%% file: sections/abstract.tex
Evaluating embodied robot foundation models remains a critical bottleneck; unlike large language models efficiently assessed via digital benchmarks, robotic policies require slow, costly real-world rollouts limited by hardware and human supervision, which has driven interest in world models as surrogate policy evaluators, yet the key properties that make a world model reliable for policy assessment remain poorly understood. This work presents a systematic study of world models for robotic policy evaluation and introduces \textit{WMBench}, a benchmark constructed from real-robot teleoperation data and matched policy rollouts covering diverse manipulation tasks to enable controlled comparisons across model families, action encodings, rollout horizons, and evaluation metrics. Using WMBench, we analyze 7 video world models, 4 action representation schemes, and over 324,000 simulated policy rollouts paired with real robot executions, further enriching our analysis with large-scale community submissions from the CVPR 2026 GigaBrain Challenge, curated synthetic trajectories, and a training videos spanning more than 12,000 hours. Our experiments deliver three core insights: evaluator quality is dominated by long-horizon, action-faithful rollout consistency rather than short-term visual realism; pretraining gains stem not only from data scale but from balancing general world knowledge with robot-specific controllability; and architectural choices including action encoding, memory design, and evaluator-focused post-training strongly determine alignment with real-world robot behavior. Drawing on these results, we derive a practical design roadmap and realize it in \textit{GigaWorld-1}, a world model specially optimized for policy evaluation, and we fully release our code, models, datasets, and toolkits to advance scalable evaluation research for embodied foundation models.

%% file: sections/introduction.tex
\section{Introduction}
Efficient evaluation is critical to the iterative improvement and performance tuning of large foundation models. In language modeling, evaluation incurs relatively low overhead: newly saved model checkpoints can be rapidly assessed against standardized benchmarks ~\citep{hendrycksmeasuring,wang2024mmlu,zhong2024agieval,lin2022truthfulqa,cobbe2021training,rein2023gpqa,jimenez2024swe,hsieh2024ruler,yue2024mmmu,yue2025mmmu,fu2025video}, meaning evaluation rarely becomes a bottleneck. Robotics, however, presents a fundamentally distinct scenario. Validating a robot policy typically demands repeated real-world rollouts on physical robot hardware, which necessitates continuous human monitoring and occupies robot hardware for lengthy evaluation cycles. As a result, evaluation emerges as the primary bottleneck holding back progress in robot policy models.

This bottleneck is especially severe for recent robot foundation models such as vision-language-action models and world-action models~\citep{liu2026towards,pi_0,pi05,gr00t,worldvla,motus,cosmospolicy,shen2025videovla,swiftvla,walloss,galaxea,team2026gigabrain05,gigaworldpolicy}. Although these models are becoming increasingly capable, reliable evaluation still depends heavily on real-robot execution. As reported in OpenVLA~\citep{kim2024openvla}, it requires 100 human hours for 2,500 rollouts evaluation, and physical robots cannot yield fully consistent reset states across trials.
Classical simulation can partially reduce cost, but its utility is limited by the sim-to-real gap, alongside prohibitive overhead associated with scene-wise digital twin construction~\citep{robotwin,robocasa,libero,simplerenv,vlabench,metaworld,rlbench,robosuite,wonderturbo,ni2025wonderfree,wang2025drivegen3d}. World models offer a compelling middle ground. Recent progress in video generation and action-conditioned world modeling~\citep{svd,cogvideox,ltx,wan,enerverse,irasim,drivedreamer,drivedreamer2,embodiedreamer,gigaworld0,liu2026towards,lang2026vag} reveals that learned world models can capture rich visual dynamics and, to some extent, controllable physical evolution~\citep{abotphysworld,roboscape,physctrl,physgen,physdreamer,cosmos3,kang2024far}. If such models can interact with robot policies and accurately preserve the relative success or failure of their rollouts, they could serve as efficient policy evaluators, reducing dependence on repeated real-world testing. However, current literature ~\citep{he2026pre,qin2024worldsimbench,quevedo2025worldgym,tseng2025scalable,li2025worldeval,guo2025ctrl,bar2025navigation} mostly demonstrates that world models \emph{can} be used for evaluation, while leaving open the more fundamental question of which designs are reliable for building world models as policy evaluator.

This paper addresses that gap. Rather than presenting a single new evaluator and reporting one headline number, we ask a broader scientific question: \emph{what matters in building world models for evaluating robot policies?} Our aim is to move the field from proof-of-concept demonstrations toward principled design rules. We focus on three concrete questions. \textbf{First}, how should one systematically evaluate whether a world model is a good policy evaluator, beyond generic video quality metrics? \textbf{Second}, how do pretraining and training data affect evaluator quality? \textbf{Third}, which architectural and algorithmic design choices most strongly influence evaluator reliability?

To answer these questions, we construct \textit{WMBench}, a benchmark centered on paired real-world and world-model rollouts. The benchmark covers eight task families, including rigid and deformable manipulation, and contains both teleoperated expert data and policy rollout data collected from multiple policy checkpoints. This design enables us to measure not only whether a world model generates plausible videos, but whether it preserves the relative outcomes of policies seen in the real world. On top of \textit{WMBench}, we perform a large-scale controlled study over 7 world models, 4 action representation schemes, 324,000+ world model rollouts, and a diverse set of evaluator metrics. Notably, the study is augmented with community submissions from a public challenge associated with CVPR 2026\footnote{\url{https://gigaai-research.github.io/GigaBrain-Challenge-2026/}}, which broadens the model design space beyond our in-house variants.

Our analysis leads to a clear picture. The best world-model evaluators are not simply the models with the most photorealistic frames. They are the models that remain action-faithful over long horizons, preserve pretrained world knowledge under robot-domain adaptation, and expose architectural pathways for stable iterative rollout. Drawing on these key insights, we propose \textit{GigaWorld-1}, which formalize a roadmap spanning data curation, world model training, and downstream policy evaluation. Empirically, \textit{GigaWorld-1} boosts evaluator-alignment metrics by 14.9\% compared to competitive state-of-the-art baselines. To facilitate follow-up investigations into scalable evaluation pipelines for embodied foundation models, we fully open-source our code, pre-trained model checkpoints, curated datasets, and auxiliary toolkits.

The contributions of this paper are four-folds:
\begin{itemize}[leftmargin=*,itemsep=2pt,topsep=2pt]
    \item We formulate \emph{world model as policy evaluator} as a first-class research problem and identify the central factors that govern whether a world model can predict policy quality in a way that matches real-world execution.
    \item We introduce \textit{WMBench}, a benchmark with human teleoperation trajectories and robot policy rollout data. Our exhaustive experiments on this benchmark cover 7 distinct world models, 4 alternative action representations, 8 robotic manipulation tasks, and over 324{,}000 evaluation rollouts, from which we distill a set of critical empirical conclusions.
    \item We provide a systematic empirical study showing how evaluator reliability depends on metric design, pretraining and data composition, and architectural choices such as action representation, memory, and reinforcement-learning-based post-training.
    \item We summarize these findings into a practical design map and instantiate them in \textit{GigaWorld-1}, which is trained using over 12000 hours data. Our model outperforms strong baselines by 14.9\% on the core evaluator-alignment metric. To foster subsequent research on scalable evaluation for embodied foundation models, we fully open-source our code, pre-trained model weights, curated datasets, and auxiliary toolkits.
\end{itemize}

%% file: sections/related_work.tex
\section{Related Work}
\subsection{Video Diffusion Models as World Models}
With the rapid advancement of video generation technologies, a number of powerful foundation models~\citep{wan2025wan,hongcogvideo,yang2024cogvideox,Zeroscope,gao2025seedance,seedance2025seedance,team2025kling,Veo3,liu2026arflow,liu2025difflow3d} have made it possible to generate videos with high visual fidelity, strong perceptual quality, and extended temporal duration. 
Building upon these capabilities, controllable video generation~\citep{ma2025controllable,ma2026fastvmt,ma2025followyourmotion,ma2026group,ma2025followfaster,ma2024followyouremoji,ma2025followyourclick,ma2024followpose,ma2025followcreation,xu2025clgc,xu2026smrabooth,xu2026disco,zhang2025flexiact,worlddreamer,humandreamer,humandreamerx} enables generated videos to follow specific control conditions, thereby opening up new possibilities for downstream applications such as autonomous driving~\citep{zhu2025worldsplat,guo2026genesis,zeng2025rethinking,zhou2026toward,zhou2026xiaomi} and embodied intelligence~\citep{he2026skip,gigaworld0}. 
A world model is expected to predict future states by jointly considering the current state and control signals. 
Video generation models are therefore naturally well suited to serve as foundation models for world modeling~\citep{sun2025worldplay,helios,ye2025yan,he2025matrix,xiang2025pan,liu2025mamba4d}. 
Existing studies~\citep{yin2025slow,huang2026self,zhu2026causal,zhao2026causal} transform control conditions into manually triggered directional inputs, which stimulate the generative model’s ability to predict future states. 
By further integrating Forcing-style techniques, these approaches enable real-time, indefinitely long video generation with extremely low latency.
\subsection{World Models for Robotic Learning}
World models have emerged as a cornerstone in embodied AI and robotic learning, primarily advancing the field through four distinct paradigms~\citep{zhu2024sora}. First, acting as \emph{data engines}, world models generate large-scale, diverse synthetic data to scale up policy training~\citep{embodiedreamer,gigaworld0,mimicdreamer,emma,robotransfer,egodemogen,cosmos-transfer,drivedreamer,drivedreamer2,egovid, unidrivedreamer}. Second, functioning as \emph{policies}, they integrate predictive dynamics directly into the action-generation loop, yielding robust end-to-end controllers~\citep{gigaworldpolicy, motus, cosmos3, dreamzero, fastwam, motubrain, drivedreamer-policy,yang2026wam,li2026stableidm}. Third, serving as \emph{interaction environments}, world models provide learned, visually rich testbeds for closed-loop planning or reinforcement learning~\citep{guo2025ctrl, cosmos3, drivedreamer4d, recondreamer, recondreamer++, Recondreamer-rl,he2026skip,wu2026imac}. Fourth, operating as \emph{value critics}, world models assess observations to provide long-horizon return estimates, guiding action selection and bootstrapping reinforcement learning~\citep{cosmospolicy,team2026gigabrain05,viva2026,tau0wm,aim2026}.

However, these paradigms do not directly answer whether a world model can serve as a reliable \emph{policy evaluator}: an external mechanism for judging whether a policy would successfully complete a task under realistic visual and physical dynamics. 
This distinction motivates a closer look at policy evaluation itself.

\subsection{Robot Policy Evaluation}
The evaluation of robotic policies has traditionally relied on a spectrum from real-world testing to simulation-based benchmarking. At one end, \emph{real-world evaluation} provides the most trustworthy assessment by directly measuring execution in target environments~\citep{robochallenge, roboarena}. However, real-robot testing is notoriously expensive, slow, and difficult to scale comprehensively under broad distribution shifts. To address this scalability bottleneck, \emph{traditional simulators} offer cheap, repeatable, and safe testing environments, underpinning numerous standard benchmarks~\citep{robotwin, robomemarena, robocasa, robocasa365, libero}. Yet, classical simulation often struggles with the sim-to-real gap, particularly in visually complex, contact-rich, or deformable scenarios.

Recently, utilizing \emph{world models as policy evaluators} has emerged as a compelling alternative to bridge this gap~\citep{guo2025ctrl, cosmos3}. By operating directly in visually open environments and leveraging large-scale spatiotemporal priors, world-model-based evaluators offer the scalability of simulation alongside a higher degree of visual and physical realism. They provide the distinct advantage of scalable, dynamic, and realistic evaluation without the manual modeling effort required by traditional simulators. Rather than asking whether learned evaluation is possible in principle, our study systematically asks what properties make such world-model-based evaluation trustworthy, scalable, and aligned with real-robot outcomes.

%% file: sections/preliminaries.tex
\section{Preliminaries}

\begin{figure*}[!t]
\centering
\captionsetup{type=figure, justification=justified, singlelinecheck=false}
\includegraphics[width=1\textwidth]{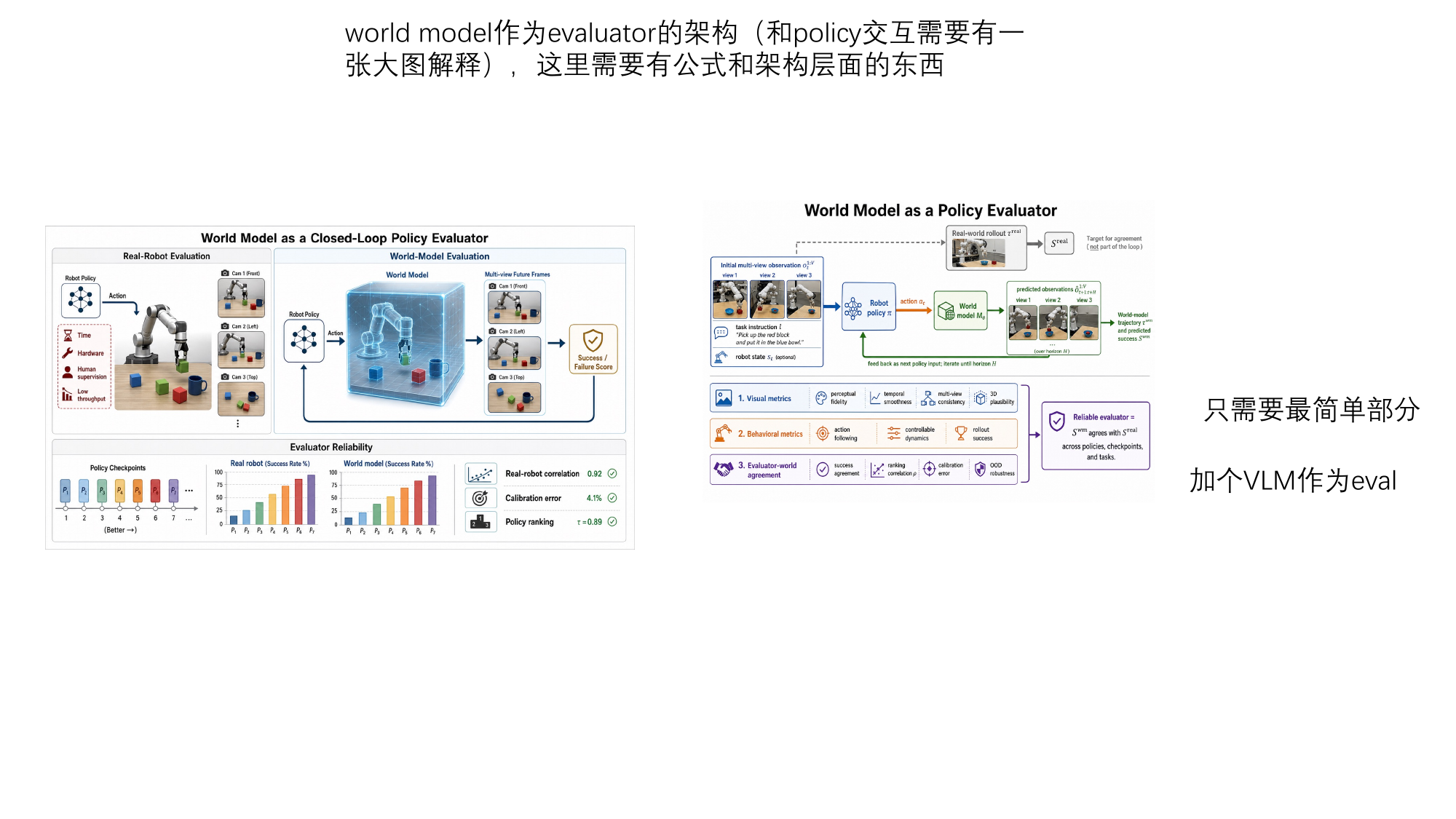}
\caption{\textbf{World model as policy evaluator framework.} A world model serves as a policy evaluator by iteratively receiving policy actions and predicting future observations. Reliable evaluation requires not only visual quality, but also action-faithful rollout and agreement with real-world policy outcomes.}
\label{fig:preliminary}
\end{figure*}

% \subsection{World model as a policy evaluator}
We consider a robot policy $\pi$ that receives an observation $o_t$, optional robot state $s_t$, and a task instruction $l$, and outputs an action $a_t = \pi(o_t, s_t, l)$. In real-world evaluation, the policy interacts with the physical environment and yields a trajectory:
\begin{equation}
\tau^{\mathrm{real}} = \{(o_t, s_t, a_t)\}_{t=1}^{T},
\end{equation}
from which we can estimate task success, failure modes, and other performance signals.

When a world model $M_\theta$ is used as an evaluator, the policy instead interacts with a learned environment. Given the initial observation, task instruction, and optionally the robot state, the model predicts future observations conditioned on the policy actions:
\begin{equation}
\hat{o}_{t+1:t+H} \sim M_\theta(\cdot \mid o_{\le t}, s_{\le t}, a_{\le t}, l),
\label{eq:wm_rollout}
\end{equation}
where $H$ denotes the rollout horizon. Iterating this process yields a world-model trajectory:
\begin{equation}
\tau^{\mathrm{wm}} = \{(\hat{o}_t, s_t, a_t)\}_{t=1}^{H}.
\end{equation}
The role of the evaluator is not merely to generate visually plausible observations; it is to preserve the decision-relevant properties of the real trajectory. In particular, for a policy set $\{\pi_i\}_{i=1}^{N}$, we care about whether world-model-based scores preserve the ranking, success prediction, and risk profile observed in the real world.

We therefore define the primary evaluator target as agreement between world-model and real-world policy outcomes. Let $S^{\mathrm{real}}(\pi)$ be the empirical success rate of policy $\pi$ in real rollouts, and let $S^{\mathrm{wm}}(\pi)$ be the predicted success rate inferred from world-model rollouts. A key alignment measure is the ranking correlation:
\begin{equation}
\rho = \mathrm{Corr}\!\left(S^{\mathrm{real}}(\pi), S^{\mathrm{wm}}(\pi)\right),
\label{eq:rank_corr}
\end{equation}
evaluated across policies, checkpoints, tasks, or rollout conditions. Throughout the paper, this evaluator-world agreement serves as the central quantity of interest.

%% file: sections/wmbench.tex
\section{WMBench: A Benchmark for World Models as Policy Evaluators}
To rigorously assess whether a world model can faithfully replace real-world physical execution for policy evaluation, we introduce WMBench. This section details its construction logic: we first describe the dataset composition and processing pipelines (Sec.~\ref{sec:wmbench_data}), then outline the four-step closed-loop evaluation protocol (Sec.~\ref{sec:wmbench_protocol}), and finally define the hierarchical metric system used to measure both generation quality and evaluator reliability (Sec.~\ref{sec:wmbench_metrics}).

\subsection{Data Source}
\label{sec:wmbench_data}

WMBench is constructed to bridge the gap between visual generation and physical policy evaluation. It also serves as the official benchmark for the GigaBrain Challenge @ CVPR 2026 World Model Track, with its dataset open-sourced on Hugging Face\footnote{\url{https://huggingface.co/datasets/open-gigaai/CVPR-2026-WorldModel-Track-Dataset}}, accumulating over 50,000 downloads.

\textbf{Data source.} The benchmark corpus comprises 2{,}989 paired trajectories across eight tasks, drawn from two complementary sources: a teleoperated real-world dataset covering varied manipulations and camera views, and a policy-rollout dataset generated by GigaBrain~\citep{gigabrain0} checkpoints containing both successes and failures, the ratio of teleoperated and rollout data is near 1:1. To ensure evaluation integrity, the train-test split strictly enforces \emph{episode-disjointness} (no test trajectory overlaps with training data), \emph{diversity preservation} (stressing generalization over memorization), and \emph{outcome balance} (distinguishing visually similar success/failure cases). After filtering, this yields a final training set of 82{,}470 seconds and a test set of 7{,}200 seconds.

\textbf{Data cleaning.} We perform conservative data cleaning before benchmark release. Specifically, we systematically remove corrupted or truncated videos, clips with camera desynchronization, trajectories with missing robot states, and episodes where control timestamps cannot be aligned to observations. Furthermore, rollouts whose outcome labels are ambiguous after human verification are excluded, and near-duplicate teleoperation trajectories are collapsed to reduce redundancy and improve effective data diversity.

\textbf{Large-scale rollout dataset.} To rigorously analyze evaluator reliability (as discussed later in Sec.~\ref{sec:experiment}), we further curated a massive annotated rollout dataset. From the submissions of over 100 participating teams in the CVPR 2026 challenge, we sampled 324,000 world model rollout segments. These segments were sequentially chained together in a closed-loop manner to form complete, long-horizon policy-world model interaction episodes (with each complete episode comprising approximately 20 to 30 rollout segments). Ultimately, these full long-horizon interaction episodes were subjected to meticulous human annotation based on a four-level ordinal scale, which is defined as \textbf{World Model as Evaluator Score (WMES)}:

\begin{itemize}[leftmargin=*,nosep]
    \item \textbf{Score 3 (accurate outcome \& high fidelity)}: The generated rollout successfully predicts the same task outcome (success/failure) as the real-world reference. Additionally, the visual execution is of high quality: the robot action and object states perfectly align with the reference, exhibiting no obvious object distortion and maintaining realistic physics and collision dynamics.
    \item \textbf{Score 2 (accurate outcome \& low fidelity)}: The generated rollout predicts the correct final outcome, but the intermediate visual generation is flawed. The process may suffer from noticeable object distortion, unrealistic physics/collisions, or slight misalignment in the robot's action trajectory.
    \item \textbf{Score 1 (incorrect outcome \& high fidelity)}: The generated rollout fails to predict the correct task outcome (e.g., the reference succeeds but the model predicts failure, or vice versa). However, the video itself remains visually and temporally stable, with the robot arm generally following the conditioned action even if the object interaction is incorrect.
    \item \textbf{Score 0 (incorrect outcome \& low fidelity)}: The generated rollout completely fails to match the reference outcome and suffers from severe generation collapse. Both the robot action and object state are visually unstable, highly distorted, or physically nonsensical.
\end{itemize}
To ensure fairness and consistency, each rollout was independently scored by three annotators, with a fourth senior annotator conducting random spot checks. A comprehensive evaluation rubric is available online\footnote{\url{https://gigaai-research.github.io/GigaBrain-Challenge-2026/guide/evaluation-rubric.html}}. These annotated rollouts provide the critical foundation for the experimental analysis in the subsequent sections.

\subsection{Evaluation Protocol}
\label{sec:wmbench_protocol}

\begin{figure*}[htbp]
\centering
\captionsetup{type=figure, justification=justified, singlelinecheck=false}
\includegraphics[width=0.75\textwidth]{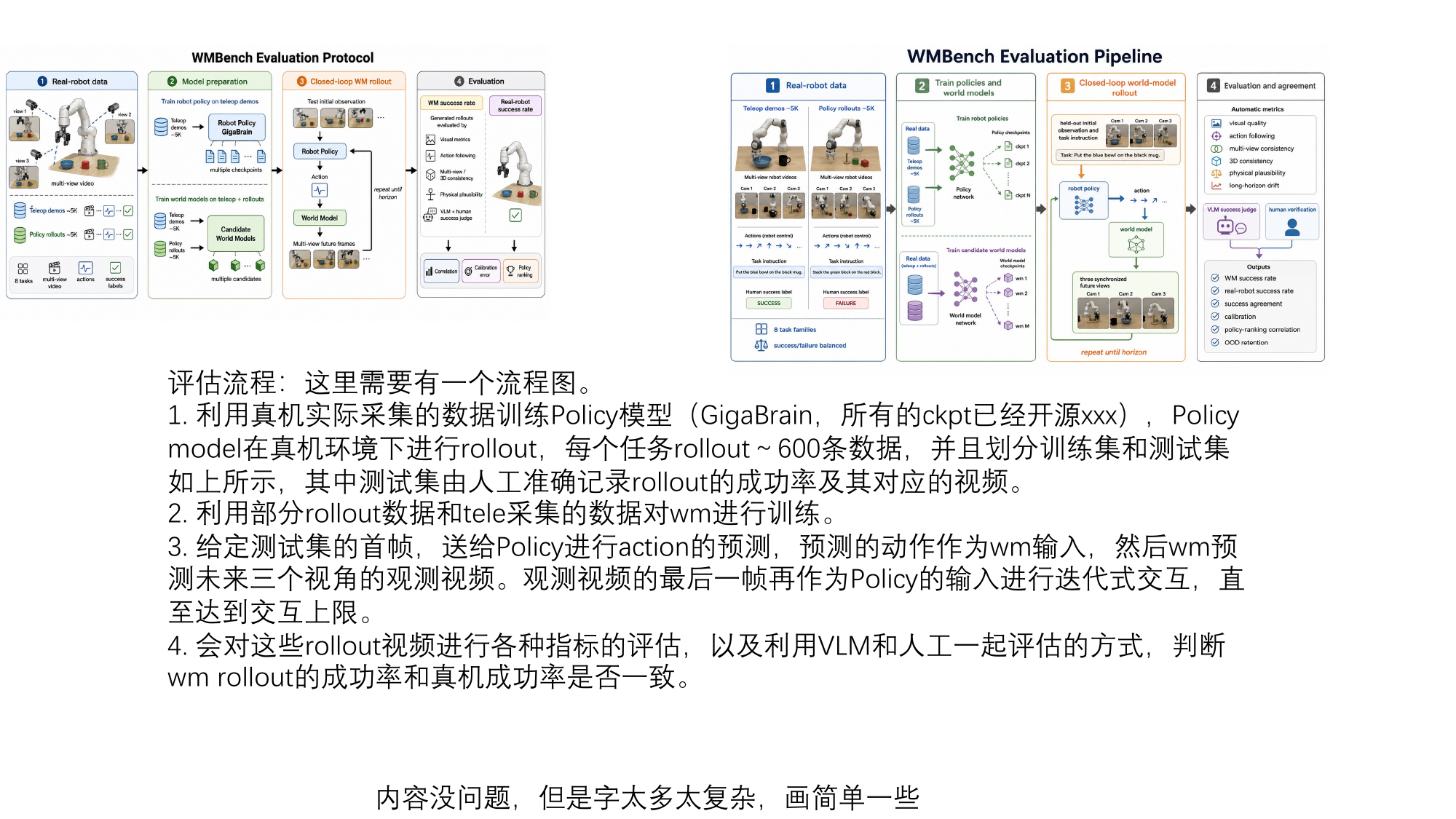}
\caption{\textbf{WMBench evaluation pipeline.} The four-step protocol includes (1) collecting real-world policy rollouts, (2) training world models on a strict split, (3) executing closed-loop policy rollouts inside the learned world model, and (4) assessing metrics and outcomes to measure alignment with real-world conclusions.}
\label{fig:wmb_pipeline}
\end{figure*}

As illustrated in Figure~\ref{fig:wmb_pipeline}, the WMBench evaluation protocol mimics the real-world deployment of a world model as a policy evaluator through four standardized steps:

\textbf{Step 1: real-world policy data collection.} We first train policy models on real-robot data and collect physical closed-loop rollouts from multiple policy checkpoints. For each episode, we record the initial observation, task instruction, multi-view rollout video, and human-annotated success labels.

\textbf{Step 2: world model training and holdout.} World models are trained on the designated training split. The test episodes, including their specific object layouts and initial states, are strictly held out to ensure the world model is evaluated on its generalization ability rather than memorization.

\textbf{Step 3: closed-loop rollout in world models.} Starting from the first frame of a held-out test episode, the target policy predicts an action. The world model then takes this action and the current state to predict future observations across synchronized views. This predicted observation is fed back to the policy for the next step, forming a closed-loop execution until task termination.

\textbf{Step 4: metric calculation and outcome assessment.} The generated trajectories are then comprehensively evaluated using the metric system defined in Sec.~\ref{sec:wmbench_metrics}. This involves two parallel streams: (1) calculating automatic metrics that capture visual fidelity and physical motion dynamics, and (2) assessing the final rollout score (WMES) via a hybrid evaluator (human annotation or VLM). This dual assessment verifies not only the generation quality but also whether the simulated outcome faithfully matches the real-world ground truth.

\subsection{Metric System}
\label{sec:wmbench_metrics}

Our metric system separates \emph{outcome evaluation} from \emph{rollout diagnostics}. The outcome-based WMES introduced above measures whether a generated rollout supports task-level decision making. For diagnostic analysis, we select a subset of automatic metrics from WorldArena~\citep{worldarena}, keeping their original definitions and evaluation protocols whenever the metric is shared. These diagnostics describe frame quality, representation fidelity, geometry, semantics, interaction, motion, and long-horizon rollout behavior. We organize them into three families: \emph{frame and representation fidelity}; \emph{geometry, semantics, and interaction}; and \emph{motion and long-horizon rollout}. Among them, Aesthetic Quality, Image Quality, JEPA Similarity, Semantic Alignment, Subject Consistency, and Trajectory Accuracy are the six diagnostics reported in the summary tables of Sec.~\ref{sec:Experiments}.

\textbf{Frame and representation fidelity.} These metrics characterize perceptual quality, content stability, and feature-level similarity:
\begin{itemize}[leftmargin=*,nosep]
    \item \emph{Image Quality and Aesthetic Quality.} Image Quality measures frame clarity and technical distortions such as blur, overexposure, noise, and compression artifacts using a MUSIQ-style no-reference image-quality predictor~\citep{musiq}. Aesthetic Quality measures visual appeal, lighting, and color composition using the LAION/CLIP aesthetic-predictor protocol~\citep{laion2022aesthetic}. We average both scores over frames.
    \item \emph{JEPA Similarity.} Generated and reference videos are encoded with a frozen V-JEPA video encoder~\citep{vjepa}, and their feature distributions are compared with polynomial-kernel MMD. Higher scores indicate closer high-level spatiotemporal alignment with reference rollouts~\citep{beyond_fvd}.

    \item \emph{Subject Consistency and Background Consistency.} Subject Consistency measures object stability across frames using DINO-style features, while Background Consistency measures scene stability using CLIP-style features~\citep{dino,clip}. Both compare the current frame with the first and previous frames and apply a dynamic-degree penalty so that near-static rollouts cannot obtain artificially high consistency scores.

    \item \emph{Photometric Consistency.} Photometric Consistency measures pixel-level texture stability with optical-flow-based average endpoint error. We report the normalized inverse score, with the same dynamic-degree adjustment used to avoid rewarding static videos.
  
\end{itemize}

\textbf{Geometry, semantics, and interaction.} These metrics quantify whether the generated rollout preserves spatial structure, task semantics, and action-conditioned effects:
\begin{itemize}[leftmargin=*,nosep]
    \item \emph{Geometry Accuracy and Perspectivity.} Geometry Accuracy compares generated and reference depth maps from a monocular depth estimator after median-based scale alignment~\citep{depth_anything_v2}. Perspectivity is a Qwen3-VL-judged~\citep{qwen3_vl} 3D-plausibility score over scale variation with depth, lighting consistency, and occlusion relationships, normalized from a Likert-scale output to $[0,1]$.

    \item \emph{Semantic Alignment and Instruction Following.} Instruction Following uses a Qwen3-VL~\citep{qwen3_vl} judge to assess whether the rollout matches the requested action type, target object, and task state. Semantic Alignment first uses Qwen2.5-VL~\citep{qwen2_5_vl} to produce structured descriptions of the generated and reference videos, then computes normalized CLIP-text similarity between the two descriptions.

    \item \emph{Interaction Quality.} Interaction Quality uses a prompted Qwen3-VL judge to evaluate robot-object contact, force transfer, friction, inertia, and boundary integrity. The 1--5 Likert score is normalized to $[0,1]$.

    \item \emph{Trajectory Accuracy.} Trajectory Accuracy evaluates whether the robot arm follows the reference execution path. Arm bounding boxes are extracted with a SAM-style segmentation/detection model~\citep{sam3}, converted into center trajectories after filtering and interpolation, and compared with the reference trajectory using normalized dynamic time warping (NDTW)~\citep{ndtw}. Higher scores indicate better spatial-temporal alignment and task-stage ordering.

\end{itemize}

\textbf{Motion and long-horizon rollout.} These metrics capture short-term motion behavior and long-horizon autoregressive degradation:
\begin{itemize}[leftmargin=*,nosep]
    
    \item \emph{Dynamic Degree and Flow Score.} Dynamic Degree and Flow Score are RAFT-based optical-flow diagnostics~\citep{raft}. Dynamic Degree focuses on the top 5\% highest-motion pixels to capture salient robot/object motion, while Flow Score averages optical-flow magnitude over all pixels and frames to capture global motion intensity.

    \item \emph{Motion Smoothness.} Motion Smoothness uses a frame-interpolation model to assess temporal coherence: intermediate frames predicted from neighboring frames are compared with real intermediate frames, with motion-aware weighting to avoid over-rewarding static backgrounds~\citep{vfimamba}. Higher scores indicate smoother and more physically coherent motion.
    
    \item \emph{PSNR, FID, and FVD.} For long-horizon rollouts, we additionally report standard reconstruction and distributional video-generation metrics: PSNR for pixel-level fidelity, FID~\citep{fid} for frame-level image distribution distance, and FVD~\citep{fvd} for video-level spatiotemporal distribution distance.

\end{itemize}

%% file: sections/what_matters.tex
\section{What Matters in Building World Models For Evaluating Robot Policies?}
\label{sec:experiment}
\subsection{Question I: How Should Evaluator Quality Be Assessed?}

If world models are to become practical evaluators, we must determine not only which metrics reliably align with WMES, but also what evaluation protocol best reflects real deployment. In this section, we answer three practical questions. \textbf{First}, leveraging the 324,000 manually annotated rollout trajectories introduced in Sec.~\ref{sec:wmbench_data}, we systematically analyze the correlation between various visual/motion metrics and WMES. \textbf{Second}, we show that evaluator quality should be judged under long-horizon and OOD rollout settings rather than by short-horizon video generation quality alone. \textbf{Third}, to overcome the bottleneck of human annotation, we introduce a scalable VLM-assisted outcome annotation pipeline, whose predictions achieve near-perfect agreement with human expert WMES rankings.

Formally, for a submitted automatic metric $m$ and the ground-truth WMES score $c$, we compute the Pearson correlation over the set of valid submissions $\Omega_m$ using pairwise deletion:
\begin{equation}
\rho(m,c)
=
\frac{
\sum_{i\in\Omega_m}(m_i-\bar{m})(c_i-\bar{c})
}{
\sqrt{\sum_{i\in\Omega_m}(m_i-\bar{m})^2}
\sqrt{\sum_{i\in\Omega_m}(c_i-\bar{c})^2}
}.
\label{eq:pearson_metric}
\end{equation}
where $m_i$ and $c_i$ are the metric value and WMES score for the $i$-th submission, respectively, and $\bar{m}$ and $\bar{c}$ are their sample means. To estimate uncertainty, we compute 95\% confidence intervals using non-parametric bootstrap resampling with $10{,}000$ iterations. A metric with an upper confidence bound below zero is considered a negative predictor, meaning higher metric scores indicate lower WMES performance.

For a metric group $\mathcal{G}$, the group-level score is the mean of its metric-level correlations:
\begin{equation}
\rho(\mathcal{G},c)
=
\frac{1}{|\mathcal{G}|}
\sum_{m\in\mathcal{G}}
\rho(m,c).
\label{eq:pearson_group}
\end{equation}
This protocol makes the evaluation target explicit: a good automatic metric should rank world models similarly to their WMES score.

\begin{figure*}[t]
\centering
\captionsetup{type=figure, justification=justified, singlelinecheck=false}
\includegraphics[width=0.95\textwidth]{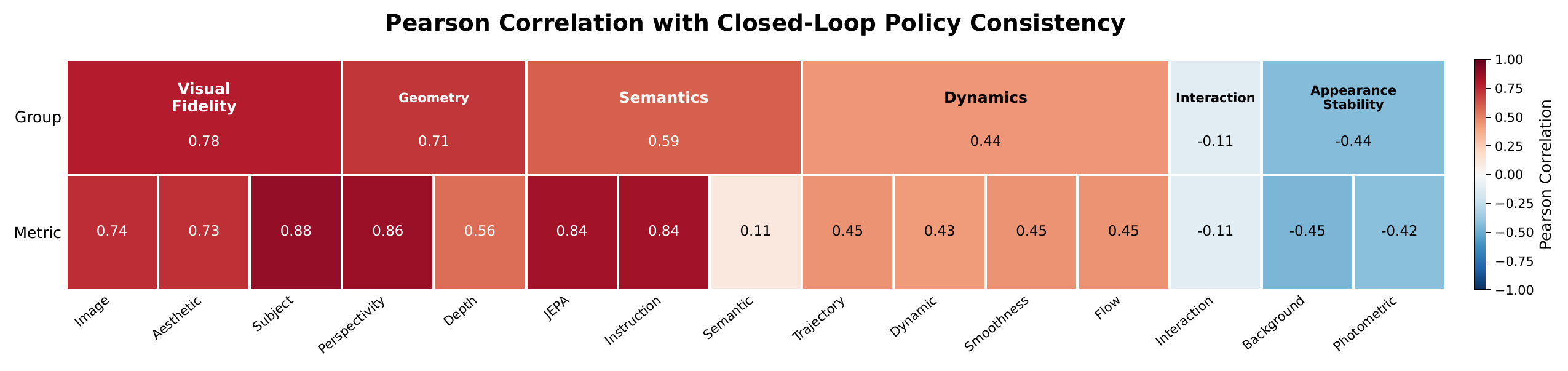}
\caption{\textbf{Metric-group correlation with WMES.} Metrics submitted to the WMBench are grouped into visual fidelity, geometry, semantics, dynamics, interaction, and appearance stability categories. Visual fidelity and geometry are the strongest group-level predictors, while appearance stability is negatively correlated with WMES.}
\label{fig:group_metric_correlation}
\end{figure*}

\begin{figure*}[t]
\centering
\captionsetup{type=figure, justification=justified, singlelinecheck=false}
\includegraphics[width=0.82\textwidth]{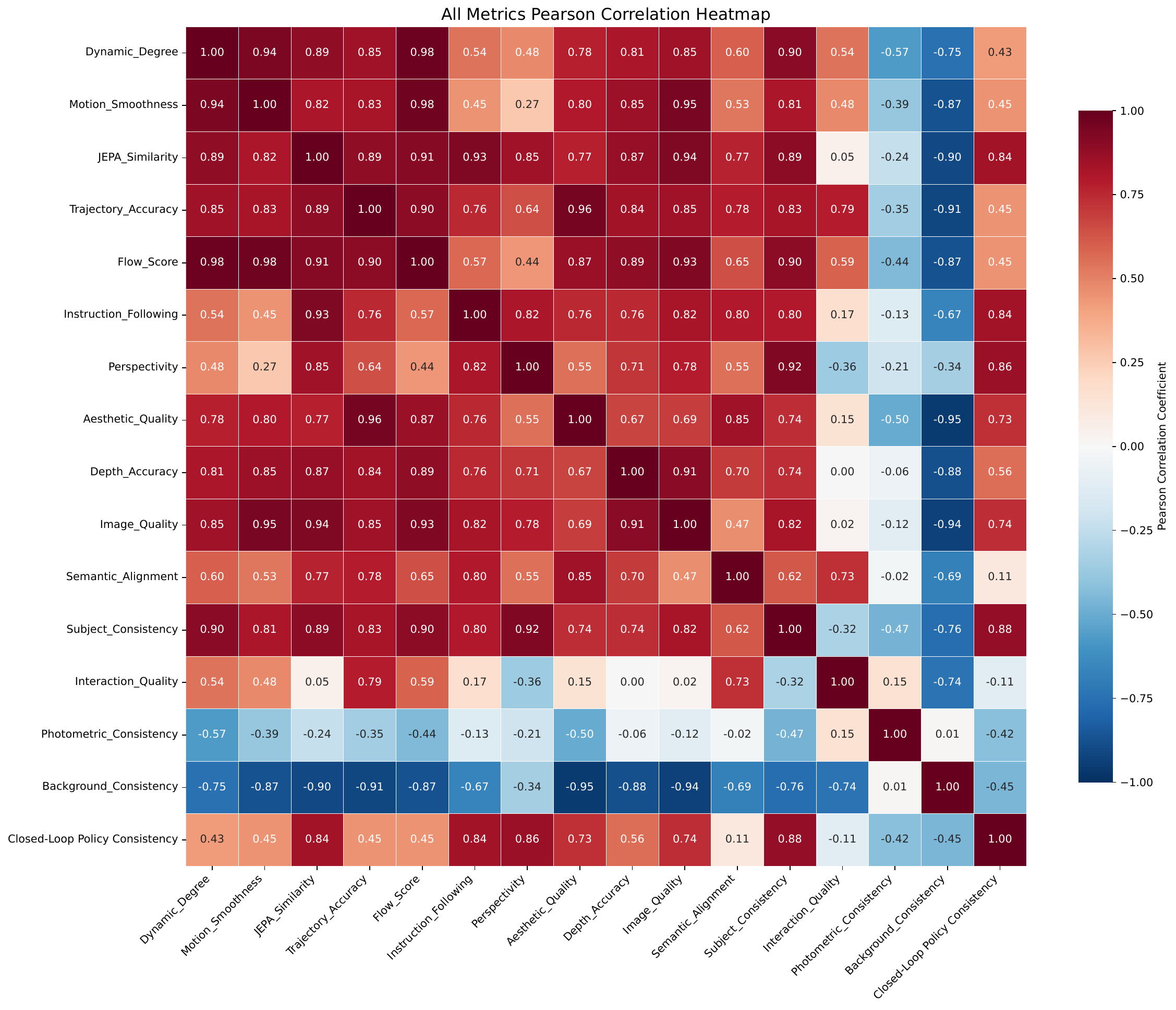}

\caption{\textbf{Pearson correlation matrix over all submitted metrics.} The full metric-level heatmap shows that Subject Consistency, Perspectivity, JEPA Similarity, Instruction Following, Image Quality, and Aesthetic Quality correlate strongly with WMES, whereas appearance-stability metrics and Interaction Quality are much less reliable.}
\label{fig:all_metric_correlation}
\end{figure*}

As shown in Figures~\ref{fig:group_metric_correlation} and~\ref{fig:all_metric_correlation}, the empirical study yields two critical findings:

\textbf{Finding 1: visual and geometric fidelity dominate WMES prediction.} 
At the group level, Visual Fidelity has the highest correlation ($\rho=0.78$), followed by Geometry ($\rho=0.71$) and Semantics ($\rho=0.59$). At the individual-metric level, Subject Consistency ($\rho=0.88$) and Perspectivity ($\rho=0.86$) are the strongest predictors, followed closely by Instruction Following ($\rho=0.84$). This suggests that a world model's ability to act as a reliable evaluator depends primarily on preserving recognizable subjects, viewpoint geometry, and semantic task information, rather than merely generating smooth motion. By contrast, Semantic Alignment alone has a weak correlation ($\rho=0.11$), showing that high-level semantic labels are insufficient if they do not capture whether the rollout preserves the policy-relevant geometric state.

\textbf{Finding 2: degenerate metrics mislead evaluator ranking.} 
Surprisingly, we identify several metrics that correlate negatively with WMES, including Background Consistency ($\rho=-0.45$), Photometric Consistency ($\rho=-0.42$), and Interaction Quality ($\rho=-0.11$). The first two fail because a trivial baseline that generates a completely static video can achieve high appearance stability while ignoring all actions and therefore failing entirely at policy evaluation. Interaction Quality is also unreliable because it is obtained by querying a VLM about physical consistency, and current VLMs still cannot judge physical realism robustly enough for fine-grained evaluator ranking. Metrics that do not explicitly penalize action-ignorance or cannot faithfully assess physical consistency can therefore favor degenerate world models over those that correctly model physical interaction.

These correlation results identify which automatic metrics are predictive, but evaluator quality still cannot be reduced to snapshot fidelity alone. Two additional findings emerge when we examine iterative rollout quality and robustness under distribution shift.

\textbf{Finding 3: evaluator quality must be assessed under long-horizon rollout, not single-step video generation.}
In practical use, world models are rolled out autoregressively, so small state errors compound over time and can eventually change the policy-level conclusion. We therefore assess long-horizon quality chunk-by-chunk over 40 seconds using PSNR for multi-view reconstruction and FID/FVD for perceptual and temporal quality. As shown in Table~\ref{tab:rollout_quality_exp}, strong short-horizon video generators do not necessarily remain strong evaluators over long rollout horizons. Generic backbones such as Wan, Cosmos, LTX, and SVD often begin with plausible short segments but then suffer from viewpoint drift, object-identity collapse, and texture accumulation, with SVD showing especially severe late-stage degradation. This shows that evaluator quality should be judged by whether a model can preserve actionable state information under repeated autoregressive feedback, rather than by one-step video-generation quality alone.

\begin{figure}[t]
    \centering
\captionsetup{type=figure, justification=justified, singlelinecheck=false}
    \includegraphics[width=1\linewidth]{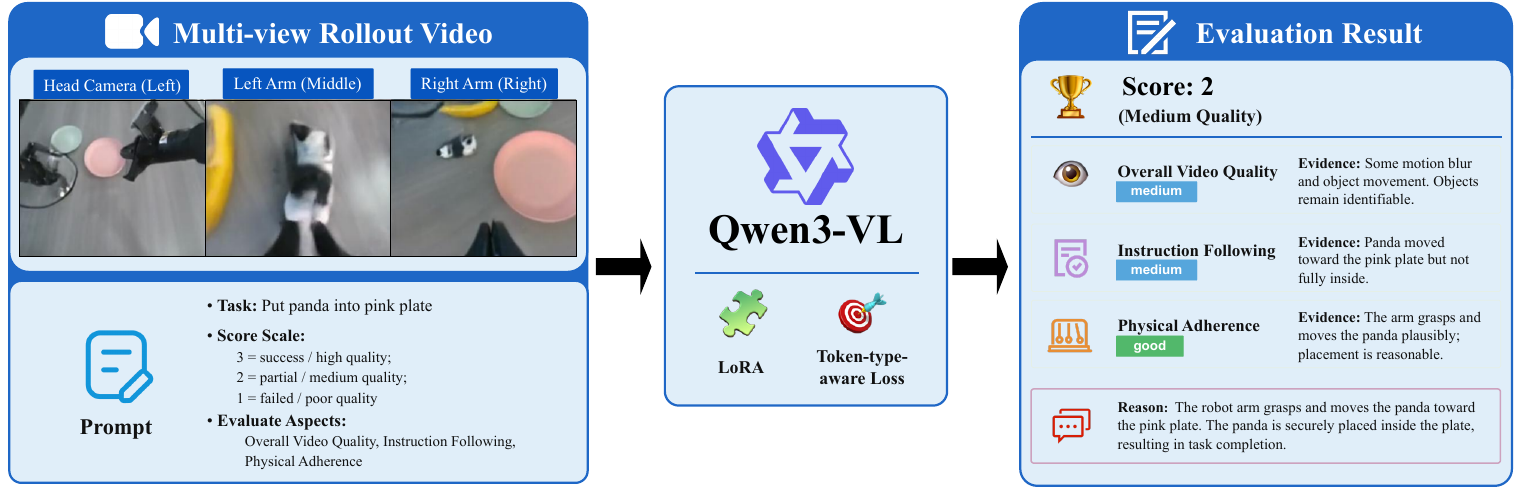}
    \caption{\textbf{VLM-assisted Rollout Evaluator.} 
    Given a three-view rollout video and a task-specific evaluation prompt, 
    the LoRA-tuned Qwen3-VL evaluator predicts a WMES score and produces 
    evidence-grounded rationales with structured aspect-level assessments of 
    \emph{overall video quality}, \emph{instruction following}, and \emph{physical adherence}.}
    \label{fig:vlm_rate}
\end{figure}

To scale outcome annotation beyond exhaustive human inspection, we train a VLM evaluator on generated rollout videos. As illustrated in Figure~\ref{fig:vlm_rate}, each rollout is presented as a synchronized three-view video together with the task-specific prompt, and the evaluator predicts both an ordinal outcome score and structured aspect-level assessments. The core challenge is to ensure that the VLM reflects human judgement on policy outcome rather than overfitting to superficial visual artifacts.

\textbf{Finding 4: outcome-centric supervision is essential for scalable VLM evaluation.}
We fine-tune Qwen3-VL-8B-Instruct using LoRA with structured supervision that couples the overall WMES score with evidence-grounded rationales, including a concise summary rationale and aspect-level assessments of \emph{overall video quality}, \emph{instruction following}, and \emph{physical adherence}. LoRA adapters with rank 16, scaling factor 32, and dropout 0.05 are applied to the attention projection layers. Each rollout video is sampled at 2 fps, with 15--32 frames used as the visual input. 

Since rationale tokens greatly outnumber the overall score token, a standard language-modeling objective would under-emphasize the final outcome prediction. Therefore, we adopt a score-focused training objective with token-type-aware loss weighting. Specifically, the overall score token receives the highest weight of 8.0, format and structure-related tokens receive a weight of 1.0 to preserve complete and parseable outputs, and free-form evidence and rationale tokens receive lower length-adaptive weights with a minimum of 0.05 to prevent verbose explanations from dominating optimization. 

\textbf{Finding 5: VLM evaluators achieve near-perfect agreement with human ratings.}
We evaluate the reliability of VLM-based evaluation by comparing VLM-predicted scores with human ground-truth annotations, as summarized in Table~\ref{tab:score_agreement}. The VLM assigns each video a WMES on the same 0--3 ordinal scale used by human annotators, and these video-level predictions are further aggregated to compute method-level WMES. Across 5,000+ videos, the VLM achieves 87.80\% exact agreement and 99.16\% adjacent agreement, with only 0.84\% of videos differing by two score levels. It also obtains low error values, with MAE 0.1304 and RMSE 0.3836, and a quadratic weighted kappa of 0.7349, indicating strong ordinal agreement with human ratings. The predictions further correlate well with human scores, achieving Spearman correlation 0.7574 and Kendall $\tau_b$ 0.7507. These results show that VLM-based evaluation closely follows human judgment trends and provides a reliable proxy for method-level WMES comparison.

\begin{table}[t]
\centering
\resizebox{\linewidth}{!}{
\begin{tabular}{cccccccccc}
\toprule
Acc. $\uparrow$ & Adj. Acc. $\uparrow$ & Large Err. $\downarrow$ & MAE $\downarrow$ & RMSE $\downarrow$ & QWK $\uparrow$ & Spearman $\uparrow$ & Kendall $\tau_b$ $\uparrow$ & W-F1 $\uparrow$ & $|\mathrm{Bias}|$ $\downarrow$ \\
\midrule
0.8780 & 0.9916 & 0.0084 & 0.1304 & 0.3836 & 0.7349 & 0.7574 & 0.7507 & 0.8744 & 0.0455 \\
\bottomrule
\end{tabular}
}
\caption{\textbf{Overall agreement metrics on 5,000+ videos.} We compare VLM-predicted scores against human ground-truth scores. Adjacent accuracy counts predictions within one score level, while large error denotes predictions that differ by two score levels. W-F1 denotes weighted F1.}
\label{tab:score_agreement}
\end{table}

\subsection{Question II: How Do Pretraining and Training Data Matter?}
The second question concerns the origin of evaluator capability. A natural hypothesis is that larger generative models or models trained on more robot data should always be better evaluators. Our results reveal a more nuanced answer: evaluator quality depends not only on scale, but also on whether the pretrained model contains transferable physical priors and whether post-training preserves them under robot-conditioned rollout.

\textbf{Finding 6: transferable physical priors matter more than raw scale in pretraining.}
A controlled pretraining-data ablation across large video foundation models is not realistic in practice, since the original datasets and training pipelines are typically unavailable. We therefore analyze a representative set of open-source video backbones with different scales and pretraining sources. Following the evidence from Question~I, we summarize each model using the core evaluator-relevant metrics that best align with WMES, while avoiding appearance-stability metrics that can be misleading for evaluator ranking. The key question is whether evaluator quality comes primarily from parameter count, or from the compatibility between pretrained priors and robot-conditioned rollout. As summarized in Figure~\ref{fig:model_arch_compare} and Table~\ref{tab:model_arch_comparison}, the evidence favors the latter. Cosmos-Predict2.5\footnote{We also plan to evaluate Cosmos-3~\citep{cosmos3}. However, its multiview version has not yet been publicly released. We are currently in discussion with the Cosmos team, and will update the comparison once access to the latest multiview model becomes available.}~\citep{cosmosp25} is the strongest baseline (AVG 0.6123), suggesting that robotics and autonomous-driving pretraining provides useful physical priors. Among general-purpose backbones, Wan 2.2 5B~\citep{wan} is the strongest (AVG 0.5948) and still surpasses the much larger LTX 2.3 (22B, AVG 0.5775)~\citep{ltx}, while CogVideoX~\cite{cogvideox} remains intermediate (AVG 0.5620). By contrast, SVD~\citep{svd} performs worst overall (AVG 0.5569) and especially struggles on Trajectory Accuracy (0.0926), indicating weak temporal and action-related dynamics despite similar scale. These comparisons show that a larger model is not automatically a better evaluator; what matters is whether the pretrained model contains transferable physical priors that can be effectively adapted to robot-conditioned prediction.

\begin{table*}[t]
\centering
\caption{\textbf{Per-metric breakdown of training-data composition ablation.} Values show absolute scores. Delta ($\Delta$) indicates the change from the GigaData-only baseline.}
\label{tab:data_ablation_detail}
\resizebox{\textwidth}{!}{
\begin{tabular}{l|ccccccc|c}
\toprule
Data Recipe & Aesthetic & Image Quality & JEPA Sim. & Photo Cons. & Semantic & Subject & Trajectory & Average \\
\midrule
GigaData & 0.34355 & 0.6904 & \textbf{0.8141} & 0.3187 & \textbf{0.8896} & 0.7212 & \textbf{0.2566} & 0.5654 \\
\midrule
GigaData + AgiBot & \textbf{0.3802} & \textbf{0.7042} & 0.5715 & 0.6218 & 0.8710 & \textbf{0.8613} & 0.1482 & 0.5940 \\
$\Delta$ & \cellcolor{rankbest}+0.03665 & \cellcolor{rankbest}+0.0138 & \cellcolor{rankmid}-0.2426 & \cellcolor{rankbest}+0.3031 & \cellcolor{rankmid}-0.0186 & \cellcolor{rankbest}+0.1401 & \cellcolor{rankmid}-0.1084 & \cellcolor{rankbest}+0.0286 \\
\midrule
GigaData + PhysData & 0.3388 & 0.6974 & 0.7678 & \textbf{0.6261} & 0.8798 & 0.7409 & 0.2497 & \textbf{0.6144} \\
$\Delta$ & \cellcolor{rankmid}-0.00475 & \cellcolor{rankbest}+0.0070 & \cellcolor{rankmid}-0.0463 & \cellcolor{rankbest}+0.3074 & \cellcolor{rankmid}-0.0098 & \cellcolor{rankbest}+0.0197 & \cellcolor{rankmid}-0.0069 & \cellcolor{rankbest}+0.0490 \\
\bottomrule
\end{tabular}
}
\end{table*}

\textbf{Finding 7: broad physical videos provide the best overall trade-off for evaluator training.}
To study post-training data composition, we adopt Wan2.1 1.3B~\citep{wan} as a common backbone and continue training it with different data mixtures. Here, \textit{GigaData} denotes our curated Giga-collected robot demonstrations with calibrated robot trajectories and multi-view observations, while \textit{PhysData} denotes the internet and physics video corpus that provides broad visual-physical priors (see Sec.~\ref{sec:data} for the full data taxonomy). Following Question~I, we interpret this ablation mainly through the metrics that are most predictive of evaluator quality---especially subject/geometry-related fidelity and the overall retained-metric average---while treating appearance-stability signals only as auxiliary diagnostics. A \textit{GigaData-only} recipe is naturally a strong baseline because its distribution is closest to the benchmark test domain. However, relying on GigaData alone can also overfit the model to a narrow robot setting and gradually erase the broader world knowledge inherited from pretraining. This is exactly where \textit{PhysData} helps: robotic manipulation involves diverse contact patterns, object dynamics, and scene changes, and general physical videos reinforce the underlying world knowledge needed to model such interactions. As shown in Table~\ref{tab:data_ablation_detail}, \textit{GigaData + PhysData} yields the strongest overall result, improving the average score from 0.5654 to 0.6144 (+0.0490). The largest gain comes from Photometric Consistency (+0.3074), together with smaller improvements in Image Quality and Subject Consistency, indicating that broad physical videos improve realism and robustness without moving the model too far away from the target robot distribution. Although JEPA Similarity, Semantic Alignment, and Trajectory Accuracy decrease slightly, \textit{PhysData} still provides the best overall trade-off because it restores general world knowledge while preserving the task-relevant bias already present in GigaData.

\textbf{Finding 8: robot-specific data mainly improves embodiment fidelity, but introduces a sharper trade-off.}
Adding AgiBot data also improves the overall score, from 0.5654 to 0.5940 (+0.0286), but the gain is more selective. In light of Question~I, the key evidence here is not the rise of appearance-oriented metrics by itself, but the trade-off between embodiment-sensitive gains such as Subject Consistency and structurally important metrics such as JEPA Similarity and Trajectory Accuracy. AgiBot substantially improves Aesthetic Quality (+0.0367), Photometric Consistency (+0.3031), and especially Subject Consistency (+0.1401), showing that robot-specific data is valuable for embodiment and camera-view fidelity. However, its diversity is still narrower than \textit{PhysData}, and introducing demonstrations from other robot embodiments can create additional mismatch when the evaluation target is already well covered by GigaData. This is reflected in the much larger drop in JEPA Similarity (-0.2426) and Trajectory Accuracy (-0.1084), indicating that narrow robot-domain data can over-specialize the generator and weaken broader structural or motion-related generalization. In other words, AgiBot can be helpful when evaluating on closely related robot platforms, but under the current GigaData-centered setting, adding broad physical videos is the better choice. The broader lesson is that evaluator-oriented training is inherently a balancing problem: robot data is useful for embodiment refinement, but broad physical data provides a better overall trade-off for reliable policy evaluation.

\begin{table*}[t]
\centering
\caption{\textbf{Control-condition ablation.} All metrics are higher-is-better ($\uparrow$). Cell colors indicate per-metric ranking: \cellcolor{rankbest}green~= best, \cellcolor{rankmid}yellow~= second best, and \cellcolor{rankworst}red~= worst.}
\label{tab:control_ablation_exp}
\resizebox{0.95\textwidth}{!}{
\begin{tabular}{llcccccc}
\toprule
Method & Control Type & Traj. Acc. $\uparrow$ & Dynamic $\uparrow$ & Smooth $\uparrow$ & Flow $\uparrow$ & Subject $\uparrow$ & Photo. $\uparrow$ \\
\midrule
Wan 2.1 1.3B I2V & None &
\cellcolor{rankworst}0.1576 & 0.2429 & 0.4997 & 0.0971 & 0.5568 & 0.2185 \\
Wan 2.1 1.3B Control & Cross-attention &
0.1620 & \cellcolor{rankworst}0.1049 & \cellcolor{rankworst}0.4525 & \cellcolor{rankworst}0.0624 & \cellcolor{rankworst}0.3573 & \cellcolor{rankworst}0.1853 \\
Wan 2.1 1.3B Control & ControlNet &
\cellcolor{rankmid}0.2566 & \cellcolor{rankmid}0.3083 & \cellcolor{rankmid}0.5197 & \cellcolor{rankmid}0.1412 & \cellcolor{rankmid}0.7212 & \cellcolor{rankmid}0.3187 \\
Wan 2.1 1.3B Control & Channel concat &
\cellcolor{rankbest}\textbf{0.3528} & \cellcolor{rankbest}\textbf{0.3566} & \cellcolor{rankbest}\textbf{0.5747} & \cellcolor{rankbest}\textbf{0.2179} & \cellcolor{rankbest}\textbf{0.8600} & \cellcolor{rankbest}\textbf{0.3206} \\
\bottomrule
\end{tabular}
}
\end{table*}

\subsection{Question III: How Do Model Design Choices Matter?}
The third question examines how architectural choices affect evaluator reliability once the data pipeline is fixed. We focus on two design axes that directly determine whether a world model can support faithful policy assessment: the action-control interface and long-horizon memory.

\textbf{Finding 9: Action control must be injected through a spatially aligned interface.}
The interface between policy actions and the world model is central. A visually plausible rollout can still be a poor evaluator if the robot or object trajectory does not follow the intended action. We therefore compare four control interfaces: no explicit control, cross-attention control, ControlNet-style spatial control, and channel-concatenated control maps.

As shown in Table~\ref{tab:control_ablation_exp}, and consistent with Question~I, the primary judge of action control is Trajectory Accuracy, while Dynamic Degree, Motion Smoothness, Flow Score, and Subject Consistency serve as auxiliary diagnostics for whether better control also preserves coherent interaction. Cross-attention provides only a marginal improvement in Trajectory Accuracy over the image-to-video baseline (0.1620 vs. 0.1576) and degrades related motion metrics, suggesting that attention-side action tokens are easily overwhelmed by appearance and semantic tokens. ControlNet-style conditioning is stronger, improving Trajectory Accuracy to 0.2566 because the control signal enters as a spatial feature.

The strongest result comes from channel-concatenated control maps, which achieve the best score across all paired metrics, including Trajectory Accuracy (0.3528), Dynamic Degree (0.3566), Motion Smoothness (0.5747), Flow Score (0.2179), Subject Consistency (0.8600), and Photometric Consistency (0.3206). This indicates that the most reliable action representation is not merely explicit, but spatially aligned with the noisy latent from the beginning of denoising. Concretely, as detailed in Sec.~\ref{sec:model_structure}, the control is implemented as a unified pixel-aligned representation derived from calibrated robot and camera geometry: the head view uses a rendered end-effector pose map to encode manipulation intent, while the wrist views use ray maps to encode view-dependent camera motion. These view-specific controls are then encoded into a shared control latent and concatenated with the noisy video latent throughout autoregressive generation, which is particularly important in multi-view settings where camera motion and object motion are easily confounded.

\begin{table*}[t]
\centering
\caption{\textbf{Long-horizon rollout quality averaged every 8 seconds.} Each 8-second interval averages eight 10-frame chunks. Cell colors indicate per-interval ranking across models: \cellcolor{rankbest}green~= best, \cellcolor{rankmid}yellow~= second best, and \cellcolor{rankworst}red~= worst.}
\label{tab:rollout_quality_exp}
\resizebox{0.6\textwidth}{!}{
\begin{tabular}{llccccc}
\toprule
Model & Metric & 0--8s & 8--16s & 16--24s & 24--32s & 32--40s \\
\midrule
\multirow{3}{*}{SVD} & PSNR $\uparrow$ & 14.05 & \cellcolor{rankworst}8.71 & \cellcolor{rankworst}7.24 & \cellcolor{rankworst}6.82 & \cellcolor{rankworst}6.88 \\
 & FID $\downarrow$ & \cellcolor{rankmid}142.84 & \cellcolor{rankworst}381.25 & \cellcolor{rankworst}423.45 & \cellcolor{rankworst}422.32 & \cellcolor{rankworst}419.21 \\
 & FVD $\downarrow$ & 173.63 & \cellcolor{rankworst}350.12 & \cellcolor{rankworst}433.91 & \cellcolor{rankworst}444.74 & \cellcolor{rankworst}443.76 \\
\midrule
\multirow{3}{*}{Cosmos2.5} & PSNR $\uparrow$ & 13.65 & \cellcolor{rankmid}13.75 & 13.40 & 13.13 & 12.83 \\
 & FID $\downarrow$ & 235.74 & 253.58 & 253.26 & \cellcolor{rankmid}243.52 & \cellcolor{rankmid}260.84 \\
 & FVD $\downarrow$ & 203.03 & 201.03 & 242.69 & 271.12 & 289.74 \\
\midrule
\multirow{3}{*}{LTX-Video} & PSNR $\uparrow$ & \cellcolor{rankworst}13.38 & 13.67 & 12.98 & 12.82 & 12.74 \\
 & FID $\downarrow$ & \cellcolor{rankworst}257.83 & 320.51 & 293.35 & 288.67 & 295.33 \\
 & FVD $\downarrow$ & \cellcolor{rankworst}217.17 & 218.13 & 232.83 & \cellcolor{rankmid}236.50 & \cellcolor{rankmid}260.78 \\
\midrule
\multirow{3}{*}{Wan 2.2} & PSNR $\uparrow$ & 14.35 & 11.57 & 10.97 & 10.53 & 10.09 \\
 & FID $\downarrow$ & 216.02 & \cellcolor{rankmid}253.02 & \cellcolor{rankmid}245.95 & 245.64 & 266.63 \\
 & FVD $\downarrow$ & \cellcolor{rankmid}169.22 & \cellcolor{rankmid}183.10 & \cellcolor{rankmid}231.50 & 259.89 & 300.10 \\
\midrule
\multirow{3}{*}{Wan 2.1} & PSNR $\uparrow$ & \cellcolor{rankmid}14.46 & 13.63 & \cellcolor{rankmid}13.63 & \cellcolor{rankmid}13.49 & \cellcolor{rankmid}13.37 \\
 & FID $\downarrow$ & 219.67 & 344.39 & 333.66 & 326.30 & 316.77 \\
 & FVD $\downarrow$ & 197.46 & 264.74 & 284.11 & 302.71 & 320.52 \\
\midrule
\multirow{3}{*}{\textit{Wan 2.1+Mem.}} & PSNR $\uparrow$ & \cellcolor{rankbest}\textbf{19.82} & \cellcolor{rankbest}\textbf{17.01} & \cellcolor{rankbest}\textbf{17.52} & \cellcolor{rankbest}\textbf{17.51} & \cellcolor{rankbest}\textbf{17.41} \\
 & FID $\downarrow$ & \cellcolor{rankbest}\textbf{40.58} & \cellcolor{rankbest}\textbf{95.22} & \cellcolor{rankbest}\textbf{111.78} & \cellcolor{rankbest}\textbf{112.66} & \cellcolor{rankbest}\textbf{121.61} \\
 & FVD $\downarrow$ & \cellcolor{rankbest}\textbf{35.30} & \cellcolor{rankbest}\textbf{68.87} & \cellcolor{rankbest}\textbf{76.15} & \cellcolor{rankbest}\textbf{84.56} & \cellcolor{rankbest}\textbf{98.34} \\
\bottomrule
\end{tabular}
}
\end{table*}

\textbf{Finding 10: reliable evaluators require persistent memory for long-horizon rollout.}
Iterative rollout creates temporal accumulation error: each generated window becomes part of the next conditioning context, so small visual or geometric mistakes can be amplified into large state errors. This is especially harmful for policy evaluation, because a rollout may look plausible in the first few frames but produce the wrong policy conclusion once object identity, camera geometry, or contact state drifts. We therefore evaluate long-horizon rollout quality chunk by chunk over 40 seconds, using PSNR for multi-view reconstruction and FID/FVD for perceptual and temporal quality. This metric choice directly follows Finding~3: once the evaluation target becomes repeated autoregressive rollout rather than single-step generation, temporal and long-horizon fidelity become the relevant evidence. As shown in Table~\ref{tab:rollout_quality_exp}, adding memory on top of the Wan 2.1 1.3B backbone substantially improves long-horizon rollout quality across all intervals. Concretely, the memory is implemented as a hierarchical history buffer with a persistent first-frame anchor and short-, mid-, and long-range temporal memories, so the model retains both the original scene identity and recent motion context; the full implementation is described in Sec.~\ref{sec:model_structure}. The long-horizon comparison also reveals several failure modes in memory-free or generic video models, including viewpoint drift, object-identity collapse, exposure artifacts, and eventual degeneration into visually meaningless frames. Qualitatively, memory reduces abrupt background jumps and preserves long-term scene identity; without it, the rollout may begin plausibly but gradually lose the state information needed for correct policy evaluation. These results show that short-horizon video quality can be misleading, and that persistent memory is necessary for reliable long-horizon evaluation.

%% file: sections/design_map.tex
\section{Final Design Map and GigaWorld-1}
\captionsetup[figure]{justification=raggedright,singlelinecheck=false}
The empirical study above can be summarized as a data-to-model-to-evaluation design map for world models as policy evaluators. At the \textbf{data level}, evaluator quality depends on balancing generic world knowledge with robot-specific controllability. At the \textbf{model level}, the best designs expose explicit low-level action representation, preserve spatial alignment, and include memory to stabilize long-horizon rollout. At the \textbf{evaluation level}, the decisive target is not visual realism in isolation, but agreement with real-world policy success under both in-distribution and OOD conditions.

We instantiate these principles in \textit{GigaWorld-1}, our final evaluator-oriented world model. As shown in Table~\ref{tab:con}, \textit{GigaWorld-1} is built from Wan~\cite{wan} backbones of two scales, \textbf{[1.3B]} and \textbf{[5B]}, and uses a training corpus composed of real robot trajectories, policy rollout data, egocentric videos, simulation-derived data, and challenge-submitted generated rollouts after quality filtering. The comparisons in Sec.~\ref{sec:experiment} suggest that both Cosmos and Wan provide strong pretrained priors for evaluator-oriented world modeling. We ultimately adopt Wan as the backbone because it is the strongest general-purpose open backbone in our controlled comparison, while also offering a more mature ecosystem for redesign and engineering. Data cleaning removes corrupted and duplicated clips, while data labeling includes success annotation, action synchronization, and quality control for multi-view consistency. On the modeling side, \textit{GigaWorld-1} combines the recommended explicit control interface, hierarchical memory, relative temporal encoding, and a progressive multi-stage training pipeline to improve evaluator stability under long-horizon rollout.

\begin{table*}[t]
\centering
\caption{\textbf{Design map of \textit{GigaWorld-1}.}}
\label{tab:gigaworld1}
\resizebox{\textwidth}{!}{%
\begin{tabular}{lcc}
\toprule
Component & Design choice & Notes \\
\midrule
Backbone & Wan-[1.3B / 5B] & open baseline with competitive performance and a mature ecosystem \\
Training data & physical videos + open-source robot + egocentric + Giga-collected data & broad physical priors with embodiment diversity \\
Data curation & quality + motion + distribution filtering & removes noisy, static, and misaligned samples \\
Structured supervision & semantic masks + depth + fast-slow captions & improves geometry and task grounding \\
Action interface & explicit pixel-aligned representation & EE pose maps + ray maps with released calibration toolkit \\
Long-horizon module & memory-augmented rollout & first-frame anchor + hierarchical history \\
Temporal encoding & Relative RoPE & reduces position drift in long autoregressive rollout \\
Training recipe & progressive multi-stage training & foundation pretraining + AR learning + optional scene LoRA + distillation \\
% Post-training & improves outcome agreement \\
\bottomrule
\end{tabular}
}
\label{tab:con}
\end{table*}

\subsection{Data Sources and Data Curation}
\label{sec:data}

\begin{figure}[t]
\centering
\includegraphics[width=\columnwidth]{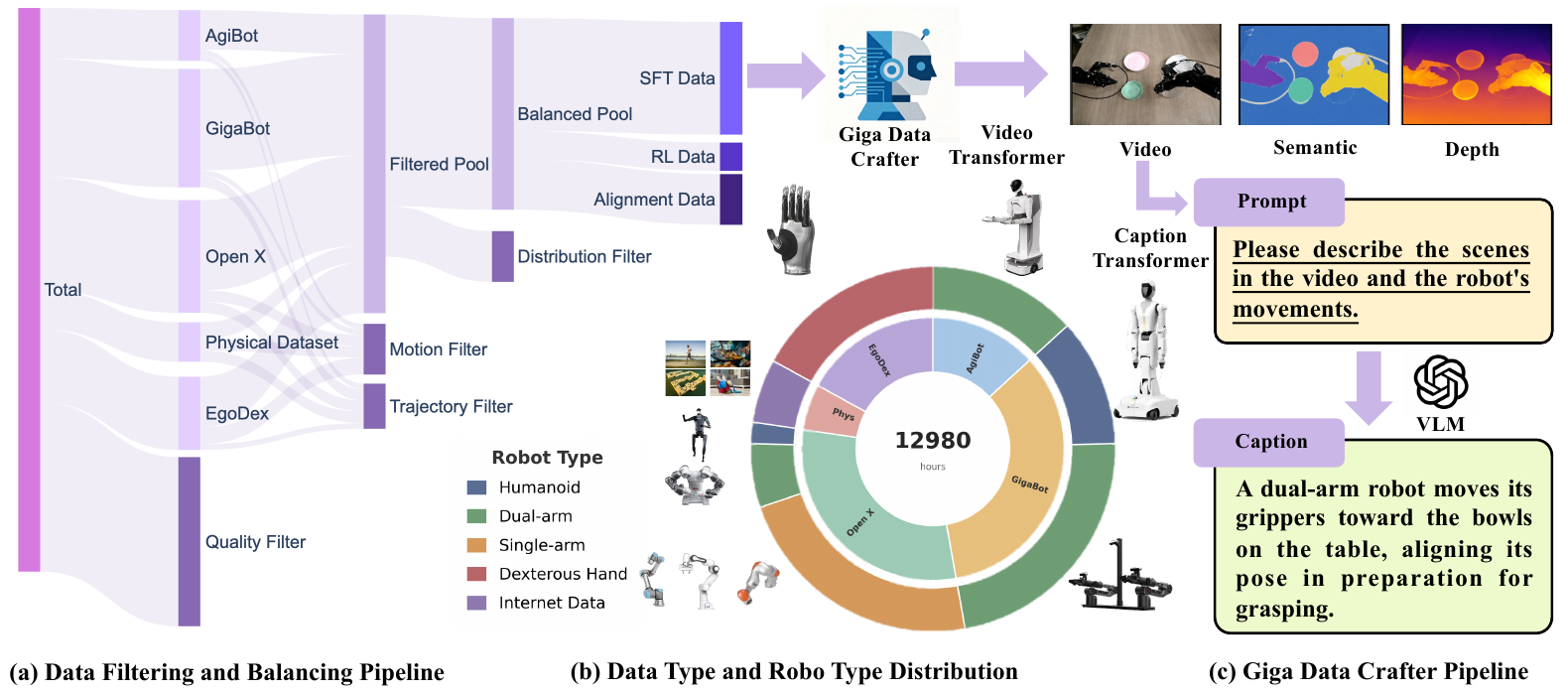}
\caption{\textbf{Data construction pipeline.} Multi-source data is filtered, balanced, and automatically annotated before being incorporated into the world-model training corpus.}
\label{fig:data_pipeline}
\end{figure}

\refstepcounter{paragraph}\paragraph*{\thesubsection.\arabic{paragraph}\quad Data Composition}
The success of large-scale embodied foundation models relies heavily on the quality, diversity, and scale of training data. The ablation results in Sec.~\ref{sec:experiment} show that broad physical-video priors are especially helpful for evaluator training: adding \textit{PhysData} provides the best overall trade-off by improving realism and robustness while preserving the task-relevant bias of robot data. Guided by this finding, and further motivated by the need for broader embodiment and scene diversity, we construct a heterogeneous corpus of approximately 12,980 hours from four complementary sources: internet and physics videos, open-source robot datasets, human-centric egocentric data, and Giga-collected robot demonstrations, as illustrated in Figure~\ref{fig:data_pipeline} and summarized in Table~\ref{tab:data_source}.

Specifically, the corpus includes about 1,298 hours of internet and physics videos for generic physical dynamics, 5,377 hours of open-source robot demonstrations, 2,411 hours of egocentric human-hand data, and 3,894 hours of Giga-collected humanoid and dual-arm demonstrations. This composition covers both broad visual-physical priors and embodiment-specific manipulation behaviors across humanoid robots, dual-arm manipulators, single-arm manipulators, and dexterous hands.

\begin{table}[t]
\centering
\caption{\textbf{Overview of the training corpus.}}
\label{tab:data_source}
\resizebox{\columnwidth}{!}{
\begin{tabular}{lccccc}
\toprule
Category &
Representative Sources &
Views &
Robot Type &
Hours &
Modality \\
\midrule

Physical Data &
Internet Videos, Physics Videos &
Single view &
N/A &
$\sim$1298 &
RGB Video \\

Open-source Robot Data &
Open X, AgiBot &
Single \& Multi Views &
Single-arm, Dual-arm and Humanoid &
$\sim$5377 &
Robot Demonstration \\

Human-centric Data &
EgoDex, SynData &
Ego-centric &
Human Hands &
$\sim$2411 &
RGB + Hand Pose \\

Giga-collected Data &
Giga Humanoid, Giga Dual-arm &
Single \& Multi Views &
Humanoid, Dual-arm &
$\sim$3894 &
Robot Demonstration \\

\midrule
Total &
Mixed &
Mixed &
Multi Types &
$\sim$12,980 &
Multi-modal \\
\bottomrule
\end{tabular}
}
\end{table}

\refstepcounter{paragraph}\paragraph*{\thesubsection.\arabic{paragraph}\quad Data Curation}
Raw aggregation alone is insufficient for training physically consistent world models, as the four data sources above differ substantially in visual style and supervision. Internet and physics videos provide broad physical priors but contain noisy web artifacts; open-source robot datasets such as AgiBot~\citep{agibot}, RoboMind~\citep{robomind}, and Galaxea~\citep{galaxea} cover diverse embodiments but use inconsistent camera and action formats; egocentric human videos emphasize hand-object interaction but lack robot trajectories; and Giga-collected demonstrations provide calibrated robot trajectories. We therefore apply a video-level quality gate followed by semantic filtering to remove corrupted, low-quality, uninformative, or misaligned samples before balancing the final training mixture.

\textbf{Video quality filter.}
We formulate video quality filtering as a clip-level acceptance test over decoded frames. For a video clip $v=\{x_t\}_{t=1}^{T}$, we first verify metadata consistency, frame decodability, timestamp monotonicity, and resolution validity; clips with codec failures, unreadable frames, duplicate-frame collapse, or abnormal aspect ratios are removed. From $K$ uniformly sampled frames, we compute a low-level image-quality vector:
  \begin{equation}
q(x_i)=\big[s(x_i), e(x_i), n(x_i), c(x_i), b(x_i)\big],
  \end{equation}
where $s,e,n,c,b$ denote sharpness, exposure validity, noise level, contrast, and compression/block-artifact scores, respectively. The aggregate image-quality score is:
  \begin{equation}
Q_{\mathrm{img}}(v)=\frac{1}{K}\sum_{i=1}^{K} w^\top q(x_i),
  \end{equation}
and clips are rejected when $Q_{\mathrm{img}}(v)<\tau_{\mathrm{img}}$, which removes videos with dirty or blurred frames, severe under/over-exposure, sensor noise, compression artifacts, or other degradations that make physical state estimation unreliable. In parallel, an aesthetic-semantic scorer $A(v)$ filters web videos with irrelevant overlays, extreme occlusion, poor framing, or non-manipulation content.

We further evaluate temporal integrity by comparing adjacent sampled frames in both histogram and embedding spaces:
  \begin{equation}
D_t=\lambda_h\bigl(1-\operatorname{sim}(h_t,h_{t+1})\bigr)
    +\lambda_\phi\bigl(1-\cos(\phi_t,\phi_{t+1})\bigr),
  \end{equation}
where $h_t$ is a color histogram and $\phi_t$ is a visual embedding. Large spikes in $D_t$ indicate scene jumps, stitching errors, black screens, or dropped frames, while persistently small changes indicate frozen or static clips. The final video-level gate is:
  \begin{equation}
\mathbbm{1}_{\mathrm{keep}}(v)
=\mathbbm{1}\!\left[
Q_{\mathrm{img}}(v)\ge\tau_{\mathrm{img}},
A(v)\ge\tau_{\mathrm{aes}},
\max_t D_t\le\tau_{\mathrm{jump}},
\operatorname{Var}_t(D_t)\ge\tau_{\mathrm{static}}
\right].
  \end{equation}
Videos that pass this gate are then constrained to training-compliant temporal windows; overly short clips are removed, while long demonstrations are segmented by task index or temporal boundaries so that each resulting clip preserves a coherent manipulation attempt.

\textbf{Motion and trajectory filter.}
After visual quality filtering, we further evaluate whether a clip contains physically meaningful manipulation dynamics. Given consecutive frames $(x_t,x_{t+1})$, we estimate a dense optical-flow field $F_t(u)\in\mathbb{R}^2$ over pixels $u\in\Omega$ and define the frame-level motion magnitude as:
  \begin{equation}
M_t=\frac{1}{|\Omega|}\sum_{u\in\Omega}\|F_t(u)\|_2 .
  \end{equation}
The clip-level kinematic score is then computed as $M(v)=\frac{1}{T-1}\sum_{t=1}^{T-1}M_t$. Clips with $M(v)<\tau_{\mathrm{motion}}$ are removed as static or uninformative, except for task segments explicitly labeled as waiting, holding, or contact stabilization. To reject physically implausible trajectories, we additionally penalize high-frequency motion artifacts using the temporal acceleration of the flow signal:
  \begin{equation}
J(v)=\frac{1}{T-2}\sum_{t=2}^{T-1}|M_{t+1}-2M_t+M_{t-1}|,
  \end{equation}
and discard clips with $J(v)>\tau_{\mathrm{jerk}}$, which captures abrupt oscillations, discontinuous camera/robot motion, and unstable hand-object contacts. For robot demonstrations, we further project calibrated action maps, including joint commands, end-effector poses, and gripper states, onto video frames. A vision-language verifier then checks whether the observed object and robot motion are consistent with the action stream, filtering samples with synchronization errors, calibration drift, or action-observation mismatches.

\textbf{Distribution filter.}
After filtering, the remaining clips are hierarchically balanced across source type, embodiment, camera view, task family, and motion intensity. The curated data is then annotated with task descriptions, success labels, action synchronization metadata, and quality tags before being used for world-model training.

\refstepcounter{paragraph}\paragraph*{\thesubsection.\arabic{paragraph}\quad Giga DataCrafter}
After curation, Giga DataCrafter converts raw videos into structured supervision for world-model training. For each retained clip $v$, the system produces three synchronized annotation streams: semantic masks, monocular depth, and language descriptions. These annotations expose object-level geometry and task semantics without requiring expensive manual labeling.

\textbf{Semantic map annotation.}
We use Segment Anything Model 2 (SAM2)~\citep{ravi2025sam} to obtain frame-level semantic masks for manipulable objects, robot arms, hands, and task-relevant background regions. Given a sampled frame $x_t$, SAM2 produces a set of masks:
  \begin{equation}
\mathcal{S}_t=\{(m_t^k,c_t^k)\}_{k=1}^{K_t},
  \end{equation}
where $m_t^k\in\{0,1\}^{H\times W}$ is the binary mask and $c_t^k$ is the corresponding semantic category or region tag. Masks are propagated and checked across neighboring frames to improve temporal consistency, yielding object-centric supervision for contact, occlusion, and spatial grounding.

\textbf{Depth annotation.}
We apply Depth Anything 3 (DA3)~\citep{lin2025depth} to estimate a dense depth map $d_t\in\mathbb{R}^{H\times W}$ for each annotated frame. The resulting depth stream provides geometric cues for hand-object distance, object support relations, and camera-view consistency. For multi-view robot data, depth predictions are further normalized per camera and checked against calibration metadata when available, producing a unified geometric signal across embodiments and viewpoints.

\textbf{Caption annotation.}
A Vision-Language Model (VLM)~\cite{Qwen3-VL,Qwen2.5-VL,Qwen2-VL,Qwen2-VL} understands video content~\cite{chen2026reflect} and generates language annotations through a fast–slow captioning system. The fast stream produces high-frequency short-term captions for local subtasks, such as reaching, grasping, lifting, placing, tool use, gripper-object contact, and object displacement. The slow stream produces low-frequency long-term captions that describe the environment in detail, including scene layout, object attributes, workspace constraints, and persistent task context. Formally, each clip is annotated as:
  \begin{equation}
\mathcal{C}(v)=\{\mathcal{C}_{\mathrm{short}}(v),\mathcal{C}_{\mathrm{long}}(v)\},
  \end{equation}
where $\mathcal{C}_{\mathrm{short}}$ is updated on short local windows and $\mathcal{C}_{\mathrm{long}}$ is updated sparsely and shared across longer temporal spans. This fast-slow design avoids repeatedly invoking large VLMs during world-model training: captions are computed offline once, cached with the video, and then reused as lightweight conditioning signals, reducing GPU cost while preserving both high-frequency subtask semantics and low-frequency environmental context.

\subsection{Architecture and Control Interface}
\label{sec:model_structure}
\setcounter{paragraph}{0}

\begin{figure*}[t]
\centering
\includegraphics[width=0.95\textwidth]{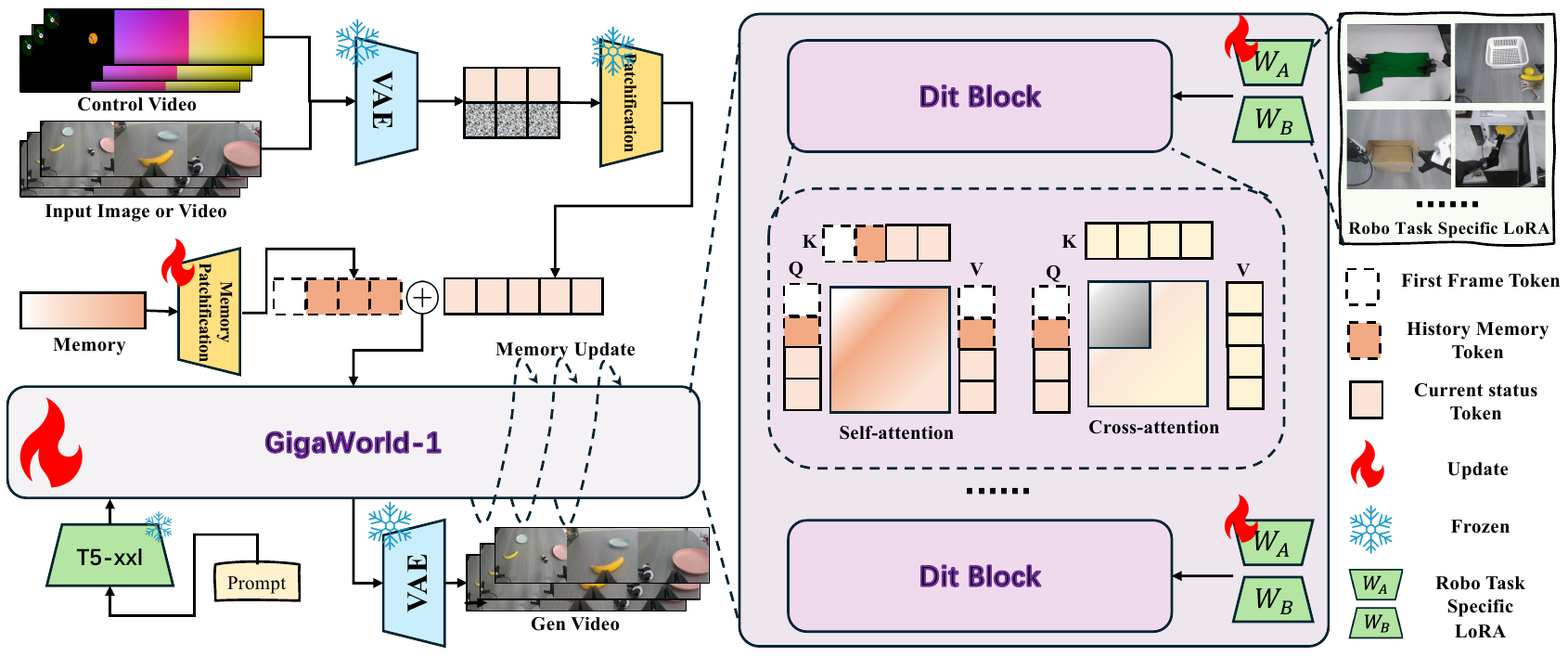}
\caption{\textbf{Overall architecture of \textit{GigaWorld-1}.} The model is built as an autoregressive diffusion-transformer world generator with parameter-efficient LoRA adaptation. Historical frames are encoded through memory patchification, future noisy latents are encoded through patchification, and structured controls such as actions, depth, semantic maps, and captions are injected as temporally aligned conditions. Frozen VAE and text-encoder components preserve pretrained visual and semantic priors, while trainable LoRA adapters and control modules adapt the DiT backbone to embodied rollout generation. Generated windows are decoded by the VAE and appended back into the rollout history for subsequent prediction.}
\label{fig:gigaworld_model_arch}
\end{figure*}

\textit{GigaWorld-1} is designed as an evaluator-oriented world model rather than a generic video generator. Its architecture follows three principles derived from the design map in Figure~\ref{fig:gigaworld_model_arch}: preserve the spatiotemporal priors of a large pretrained video backbone, inject robot actions through an explicit and geometrically aligned interface, and maintain state across iterative rollouts so that long-horizon policy outcomes remain stable. To retain pretrained generative priors while adapting efficiently to robot domains, the VAE, text encoder, and frozen backbone components are reused, while trainable LoRA adapters~\citep{hulora} and lightweight control pathways specialize the autoregressive diffusion transformer.

Figure~\ref{fig:gigaworld_model_arch} summarizes the overall architecture of \textit{GigaWorld-1}. The model extends a pretrained video diffusion backbone into an autoregressive world generator by combining memory patchification, temporally aligned control injection, hierarchical history guidance, relative temporal position encoding, and LoRA-based parameter-efficient adaptation. This design keeps the pretrained VAE and language encoder fixed, while concentrating robot-domain learning in the AR DiT adapters and control branches.

\refstepcounter{paragraph}\paragraph*{\thesubsection.\arabic{paragraph}\quad Autoregressive World Generation}
Most video diffusion backbones are trained as bidirectional denoising models over fixed-length clips, which limits their ability to synthesize long-horizon interaction trajectories. Inspired by~\citep{helios}, we reformulate world generation as an autoregressive video-continuation problem. Given a historical latent sequence:
  \begin{equation}
X_{\mathrm{hist}}=\{x_1,x_2,\ldots,x_t\},
  \end{equation}
and a future latent window:
  \begin{equation}
X_{\mathrm{future}}=\{x_{t+1},x_{t+2},\ldots,x_{t+T_f}\},
  \end{equation}
the model learns the conditional distribution:
 \begin{equation}
p(X_{\mathrm{future}}\mid X_{\mathrm{hist}}, C),
 \end{equation}
where $C$ denotes the control condition, including action maps, semantic maps, depth, captions, or task context. During training, the future window is corrupted by the diffusion process:
 \begin{equation}
X_{\tau}=\alpha_{\tau}X_{\mathrm{future}}+\sigma_{\tau}\epsilon,
\qquad \epsilon\sim\mathcal{N}(0,I),
 \end{equation}
and the network predicts the denoising target:
 \begin{equation}
\epsilon_{\theta}
=f_{\theta}(X_{\tau},H_t,C,\tau),
 \end{equation}
where $H_t$ denotes the historical memory state. During inference, each generated future segment is appended to the history buffer and used for the next prediction step, yielding the factorization:
 \begin{equation}
p(X_{1:T}\mid C)=\prod_{k=1}^{K}p(X_k\mid H_{k-1},C).
 \end{equation}
This converts a fixed-window diffusion model into an autoregressive world generator that can produce long-horizon rollouts while preserving the original denoising-based sampling procedure.

\refstepcounter{paragraph}
\paragraph*{\thesubsection.\arabic{paragraph}\quad Unified Control Injection}

To support controllable robot world generation, we introduce a unified control representation that explicitly incorporates both robot manipulation intent and camera geometry~\citep{qiu2026ge,liao2025genie}. Since different camera views exhibit distinct motion characteristics, we employ view-specific control signals while projecting them into a shared latent space.

\textbf{EE pose control.}
The head camera is approximately static with respect to the robot base. Consequently, future scene evolution is primarily determined by robot actions rather than camera motion. For this view, we adopt an end-effector (EE) pose map as the control signal. Specifically, future end-effector trajectories are projected into the image plane, producing a spatially aligned representation that encodes arm position, orientation, and gripper state~\citep{wang2025precise}. This representation provides an explicit description of future manipulation intent and enables the model to associate robot actions with subsequent scene changes~\citep{fu2025learning}.

\textbf{Ray map control.}
Unlike the head camera, wrist cameras are rigidly attached to the robot arms and move together with the end effectors. 
In this setting, most visual variations are caused by viewpoint changes rather than object motion. Therefore, we use a ray map as the control signal for wrist views~\citep{zheng2024cami2v}. 
For each pixel, the ray map stores the ray origin and normalized ray direction in the world coordinate system, providing an explicit representation of camera geometry~\citep{attal2023hyperreel}. Since the ray map changes synchronously with robot motion, it helps the model distinguish appearance changes induced by camera motion from those caused by scene dynamics.

Let \(C_t^{\mathrm{ee}}\) be the rendered EE pose map for the head view, and \(C_t^{\mathrm{ray}}\) be the ray map for the wrist view. We concatenate them along the image-width dimension to obtain a unified control map:
 \begin{equation}
C_t=\operatorname{Concat}_W(C_t^{\mathrm{ee}}, C_t^{\mathrm{ray}}),
 \end{equation}
where \(\operatorname{Concat}_W(\cdot)\) denotes width-wise concatenation. The control sequence \(C=\{C_t\}_{t=1}^{T}\) is then encoded into the latent control representation:
 \begin{equation}
Z_{\mathrm{ctrl}}=E(C).
 \end{equation}

During autoregressive generation, the \(k\)-th generation window uses the temporally aligned control segment:
 \begin{equation}
Z_{\mathrm{ctrl}}^{(k)}=Z_{\mathrm{ctrl}}[t_k:t_k+T_f],
 \end{equation}
where \(T_f\) is the temporal length of the current generation chunk. The denoising network predicts the noise conditioned on the noisy latent, the aligned control latent, the hierarchical history memory, and the diffusion timestep:
 \begin{equation}
\epsilon_{\theta}
=f_{\theta}(X_{\tau}, Z_{\mathrm{ctrl}}^{(k)}, H_t, \tau).
 \end{equation}
Unlike one-shot conditioning, this formulation continuously provides temporally aligned control signals throughout autoregressive generation. By combining EE pose control for static head views and ray map control for dynamic wrist views in a unified latent representation, the model jointly captures robot kinematics, camera geometry, and scene dynamics, enabling accurate long-horizon prediction of robot-object interactions.

\refstepcounter{paragraph}\paragraph*{\thesubsection.\arabic{paragraph}\quad Hierarchical History Injection}
Autoregressive generation must preserve both short-term motion continuity and long-term scene consistency under a bounded context budget. Instead of conditioning on all previous frames, we represent history with a multi-scale memory:
 \begin{equation}
H_t=\{H_t^{(L)},H_t^{(M)},H_t^{(S)}\},
 \end{equation}
where \(H_t^{(S)}\) captures local motion continuity and recent scene changes, \(H_t^{(M)}\) stores intermediate action evolution and object interactions, and \(H_t^{(L)}\) preserves global scene layout, object identity, and environmental state.

To mitigate identity, color, and scene drift, we additionally keep the first-frame latent \(x_{\mathrm{anchor}}\) as a persistent anchor. The final history state is:
 \begin{equation}
\widetilde{H}_t=\{x_{\mathrm{anchor}},H_t^{(L)},H_t^{(M)},H_t^{(S)}\}.
 \end{equation}
The denoising network is then conditioned on the control latent and the anchored hierarchical history:
 \begin{equation}
\epsilon_{\theta}
=f_{\theta}(X_{\tau}, Z_{\mathrm{ctrl}}^{(k)} \widetilde{H}_t, \tau).
 \end{equation}
Since the anchor is never evicted during memory updates, each generation step retains access to the original appearance statistics, while the hierarchical memory provides multi-scale temporal context. This combination improves long-horizon consistency while controlling memory consumption.

\refstepcounter{paragraph}\paragraph*{\thesubsection.\arabic{paragraph}\quad Hierarchical History Guidance Attention}

As shown in Figure~\ref{fig:gigaworld_model_arch}, the historical and noisy contexts exhibit fundamentally different statistics and should therefore be treated differently. The historical context contains previously generated observations that have already been aligned with the task objective and environmental dynamics. In contrast, the noisy context corresponds to the future latent window that requires denoising and prediction. Therefore, the role of historical context is not to be regenerated, but rather to guide the generation of future observations.

To achieve this, we maintain a hierarchical history memory consisting of a first-frame anchor and multi-scale temporal memories:
\begin{equation}
X_{\mathrm{Hist}}
=
\left\{
X_{\mathrm{Anchor}},
X_{\mathrm{Long}},
X_{\mathrm{Mid}},
X_{\mathrm{Short}}
\right\}.
\end{equation}
The anchor memory preserves the initial scene configuration and object identity, while long-, mid-, and short-term memories capture scene-level, task-level, and motion-level information, respectively.

In the self-attention layer, we compute the query, key, and value tensors for the noisy and historical contexts, denoted $Q_{\mathrm{Noisy}},K_{\mathrm{Noisy}},V_{\mathrm{Noisy}}$ and $Q_{\mathrm{Hist}},K_{\mathrm{Hist}},V_{\mathrm{Hist}}$ respectively.
The self-attention is computed as:
\begin{equation}
X_{\mathrm{Self}}
=
\operatorname{Attention}
\left(
[Q_{\mathrm{Noisy}},Q_{\mathrm{Hist}}],
[K_{\mathrm{Noisy}},K_{\mathrm{Hist}}],
[V_{\mathrm{Noisy}},V_{\mathrm{Hist}}]
\right).
\end{equation}
Unlike conventional self-attention, the hierarchical history memory serves as a persistent guidance source that provides long-term scene context, task progress, and motion continuity for future prediction.

In cross-attention, semantic information is injected through the task description. Since the historical memory has already accumulated semantic information from previous generation windows, repeatedly conditioning historical tokens on the same task description is unnecessary. Therefore, we apply cross-attention only to the current noisy window:
\begin{equation}
X_{\mathrm{Cross}}
=
\operatorname{Attention}
\left(
Q_{\mathrm{Noisy}},
K_{\mathrm{Task}},
V_{\mathrm{Task}}
\right).
\end{equation}
The final representation is obtained by combining self-attention and cross-attention outputs:
\begin{equation}
X
=
X_{\mathrm{Self}}
+
X_{\mathrm{Cross}}.
\end{equation}
\refstepcounter{paragraph}\paragraph*{\thesubsection.\arabic{paragraph}\quad Relative RoPE}
Long-horizon autoregressive generation suffers from temporal position drift when a model trained with fixed absolute temporal indices is evaluated beyond its training horizon. To avoid exposing the model to unseen positional distributions, we use Relative Rotary Position Embedding, referred to as Relative RoPE.

At each autoregressive step, we build a local temporal coordinate system for the concatenation of the historical context and the current generation window. Given a history length $T_h$ and a future window length $T_f$, the concatenated sequence has length $T_h + T_f$. We assign local temporal positions as:
 \begin{equation}
\mathcal{P}_{\mathrm{hist}}=\{0,1,\ldots,T_h-1\},
\qquad
\mathcal{P}_{\mathrm{future}}=\{T_h,T_h+1,\ldots,T_h+T_f-1\}.
 \end{equation}
Equivalently, all tokens in the current autoregressive step use local temporal positions from:
 \begin{equation}
\mathcal{P}=\{0,1,\ldots,T_h+T_f-1\}.
 \end{equation}
These local positions are reinitialized at every generation step and therefore do not depend on the absolute timestamp of the generated video. For a token located at local temporal position $p \in \mathcal{P}$, rotary embeddings are applied as:
 \begin{equation}
Q_p' = R(p)Q_p,
\qquad
K_p' = R(p)K_p,
 \end{equation}
where $R(p)$ denotes the rotary embedding operator evaluated at the local temporal position $p$. In this way, the model always observes the same range of temporal positions during both training and inference, which helps reduce repetitive motion and temporal instability in long rollouts.

\begin{figure*}[t]
\centering
\includegraphics[width=0.95\textwidth]{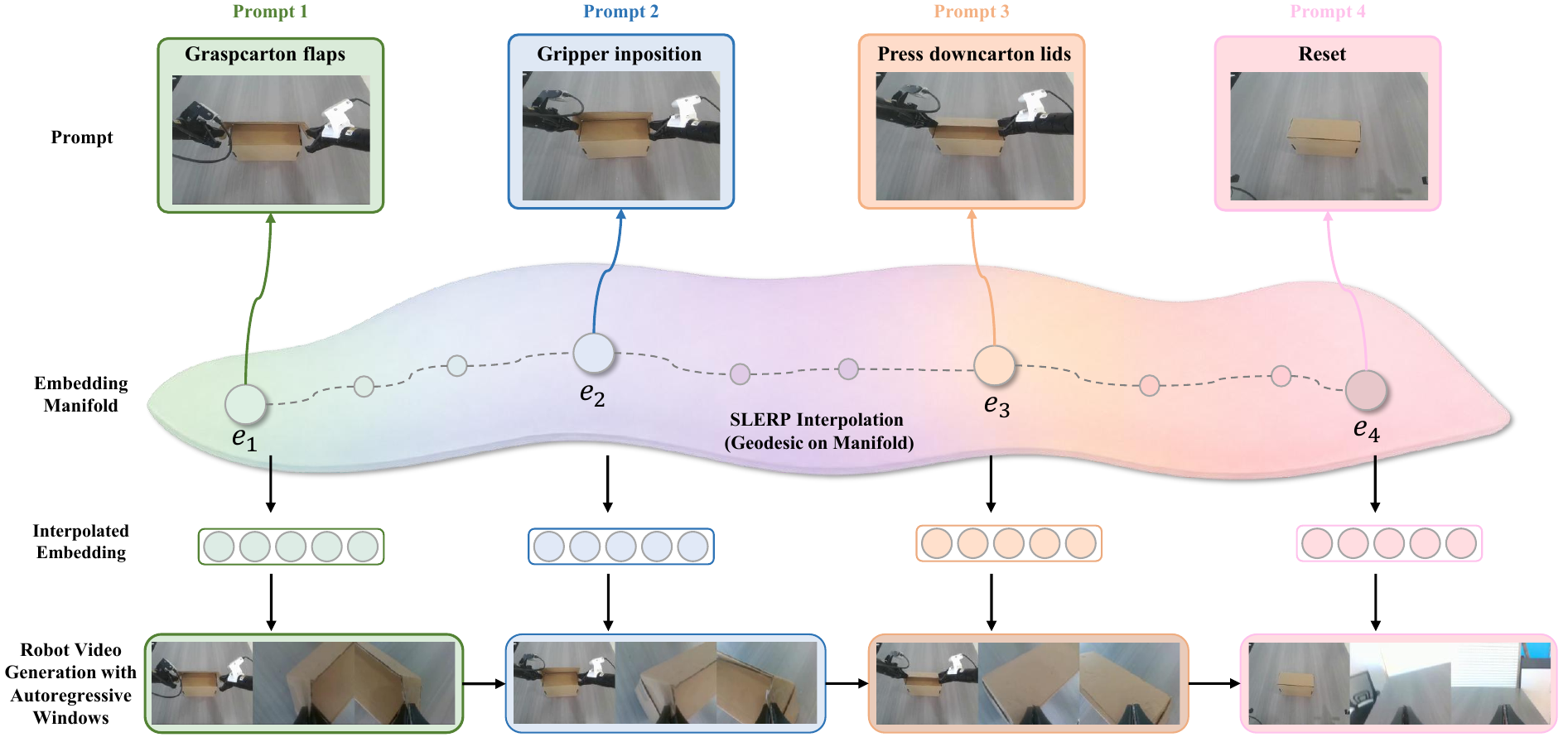}
\caption{\textbf{Prompt transition via spherical linear interpolation.} Instead of abruptly switching between prompt embeddings, \textit{GigaWorld-1} samples intermediate text conditions along the spherical path between two semantic endpoints. These interpolated embeddings are injected into successive autoregressive windows, producing smoother changes in scene appearance, motion pattern, and task phase.}
\label{fig:prompt_interpolation}
\end{figure*}

\refstepcounter{paragraph}\paragraph*{\thesubsection.\arabic{paragraph}\quad Prompt Transition via Spherical Linear Interpolation}
Long-video generation often requires smooth semantic transitions across different temporal stages. Directly switching between prompts may introduce abrupt changes in scene appearance, motion patterns, and object behaviors. As illustrated in Figure~\ref{fig:prompt_interpolation}, we achieve gradual transitions between semantic conditions by interpolating text embeddings using Spherical Linear Interpolation (SLERP).

Given two text embeddings $\mathbf{e}_1,\mathbf{e}_2 \in \mathbb{R}^{d}$ corresponding to two prompts, we first compute the angle between them:

\begin{equation}
\theta =
\arccos
\left(
\frac{
\mathbf{e}_1^\top \mathbf{e}_2
}{
\|\mathbf{e}_1\|
\|\mathbf{e}_2\|
}
\right).
\end{equation}
For an interpolation coefficient $t \in [0,1]$, the interpolated embedding is defined as
\begin{equation}
\operatorname{SLERP}
(\mathbf{e}_1,\mathbf{e}_2,t)
=
\frac{\sin((1-t)\theta)}
{\sin\theta}
\mathbf{e}_1
+
\frac{\sin(t\theta)}
{\sin\theta}
\mathbf{e}_2.
\end{equation}
To guide autoregressive long-video generation, we uniformly sample a sequence of interpolation coefficients:
\begin{equation}
t_i = \frac{i}{N-1},
\qquad
i = 0,\ldots,N-1,
\end{equation}
and obtain a sequence of text conditions
\begin{equation}
\mathbf{e}^{(i)}
=
\operatorname{SLERP}
(\mathbf{e}_1,\mathbf{e}_2,t_i).
\end{equation}
These interpolated embeddings are progressively injected into successive generation windows, allowing the world model to smoothly evolve from one semantic state to another. Compared with conventional linear interpolation, SLERP preserves the angular structure of the embedding space and better maintains semantic consistency throughout long-horizon generation.

\subsection{Progressive Training Pipeline}
\label{sec:training_pipeline}

\begin{figure*}[t]
\centering
\includegraphics[width=0.95\textwidth]{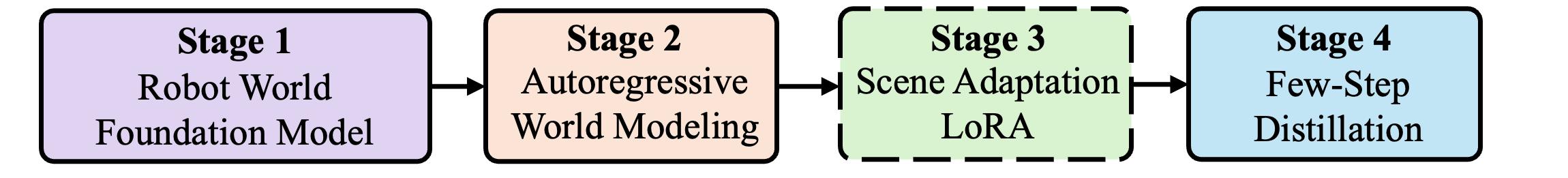}
\caption{\textbf{Training pipeline of \textit{GigaWorld-1}.} The model is first adapted into a robot world foundation model, converted into an autoregressive world generator, and finally compressed through optional ODE warm start and required DMD2 distillation for few-step rollout. Dashed branches denote optional modules that can be skipped.}
\label{fig:training_pipeline}
\end{figure*}

We train \textit{GigaWorld-1} with a progressive curriculum that separates robot-domain pretraining, autoregressive rollout learning, scene-level adaptation, and few-step distillation. As summarized in Figure~\ref{fig:training_pipeline}, the first two stages are required for obtaining a controllable long-horizon world model. Scene adaptation is optional and is used only when additional in-domain demonstrations are available. In contrast, DMD2 distillation is part of the default recipe, since evaluator-scale policy rollout requires fast generation. ODE distillation is used only as an optional warm start for DMD2: it provides a smoother initialization for the student sampler, but the final few-step model is trained by DMD2.

\textbf{Stage 1: robot world foundation model.}  We first initialize from a pretrained video backbone and continue training on the curated multi-source robot corpus. This stage learns a bidirectional robot video prior that captures embodiment-dependent kinematics, object motion, contact events, and camera-specific dynamics. Let $x_0$ denote a clean latent video, $x_t$ its noisy counterpart at continuous time $t$, and $u_t$ the target velocity. The model is optimized with the flow-matching objective:
\begin{equation}
\mathcal{L}_{\mathrm{FM}}
=
\mathbb{E}
\left[
\|
u_t-u_\theta(x_t,t)
\|_2^2
\right].
\end{equation}
where $u_\theta$ denotes the predicted velocity field. This stage does not impose autoregressive causality; instead, it transfers general spatiotemporal knowledge into the embodied domain and provides the initialization for controlled rollout learning.

\textbf{Stage 2: autoregressive world modeling.} The bidirectional foundation model is then converted into an autoregressive world model using Relative RoPE, Hierarchical History Injection, the First-Frame Anchor, and Unified Control Injection. Given historical memory $H_t$, temporally aligned controls $C$, and a noisy future latent window $X_\tau$, the model learns to denoise future observations conditioned on past context and robot actions:
\begin{equation}
\mathcal L_{\mathrm{AR}}
=
\mathbb E
\left[
\|
\epsilon-\epsilon_\theta
(X_\tau,H_t,C,\tau)
\|_2^2
\right].
\end{equation}
During training, historical frames are sampled from real trajectories; during inference, generated windows are appended to the memory buffer and reused as context for subsequent prediction. This training-inference alignment is critical for reducing compounding error in long-horizon policy evaluation.

\textbf{Stage 3: scene adaptation LoRA (optional).} For deployment in a specific workspace or robot cell, we optionally perform Low-Rank Adaptation (LoRA). The backbone parameters remain frozen and only low-rank matrices are optimized:
\begin{equation}
W=W_0+BA.
\end{equation}
This stage adapts the model to scene-specific appearance, lighting, camera calibration, object inventory, and task distribution while preserving the general dynamics learned from the full corpus. If no scene-specific data is available, the Stage-2 model is directly passed to the distillation stage.

\textbf{Stage 4: few-step distillation.} To accelerate inference, we distill the autoregressive teacher into a few-step student. ODE distillation is optional and serves as a warm-start procedure. It constructs teacher-student trajectory pairs by integrating the teacher dynamics and minimizing the discrepancy between the teacher ODE solution and the student prediction:
\begin{equation}
\mathcal L_{\mathrm{ODE}}
=
\mathbb E_{x_t,t}
\left[
\left\|
\hat{x}^{\mathrm{teacher}}_0(x_t,t)
-
\hat{x}^{\mathrm{student}}_0(x_t,t)
\right\|_2^2
\right].
\end{equation}
DMD2 is then applied as the mandatory few-step distillation objective. It combines teacher distribution matching, score consistency, and adversarial supervision so that the student remains faithful to the multi-step teacher while producing realistic rollouts with substantially fewer denoising steps:
\begin{equation}
\mathcal L_{\mathrm{DMD2}}
=
\lambda_{\mathrm{dm}}\mathcal L_{\mathrm{distill}}
+\lambda_{\mathrm{score}}\mathcal L_{\mathrm{score}}
+
\lambda_{\mathrm{GAN}}
\mathcal L_{\mathrm{GAN}}.
\end{equation}
The detailed optimization settings are reported in Tables~\ref{tab:training_hparams_pretrain} and~\ref{tab:training_hparams_distill}. Table~\ref{tab:training_hparams_pretrain} covers the non-distillation stages, while Table~\ref{tab:training_hparams_distill} isolates the optional ODE warm start and the required DMD2 distillation stage.
\begin{table*}[t]
\centering
\caption{\textbf{Detailed training hyperparameters for the non-distillation stages.} Stage is used as the horizontal axis. AdamW uses $\beta_1=0.9$, $\beta_2=0.999$, $\epsilon=10^{-8}$. This table excludes ODE and DMD2 distillation.}
\label{tab:training_hparams_pretrain}
\small
\setlength{\tabcolsep}{3pt}
\begin{tabular}{@{}p{0.26\textwidth}p{0.31\textwidth}p{0.31\textwidth}@{}}
\toprule
Configuration & Stage 1 & Stage 2 \\
\midrule
Global Batch Size & 32 & 32 \\
Optimizer & AdamW & AdamW \\
Learning Rate & $5{\times}10^{-5}$ & $1{\times}10^{-4}$ \\
Learning Rate Schedule & Cosine & Cosine \\
Training Steps & 13k & 36k \\
Gradient Clipping & 1.0 & 1.0 \\
LoRA Rank & 128 & 256 \\
LoRA Alpha & 128 & 256 \\
Numerical Precision & BFloat16 & BFloat16 \\
GPU Usage & 32 NVIDIA H20 & 32 NVIDIA H20 \\
\bottomrule
\end{tabular}
\end{table*}

\begin{table*}[t]
\centering
\caption{Detailed training hyperparameters for the distillation stages. ODE is optional and serves as a warm start, whereas DMD2 is required for the final few-step model. AdamW uses $\beta_1=0.0$, $\beta_2=0.999$, $\epsilon=10^{-8}$, and weight decay $10^{-3}$.}
\label{tab:training_hparams_distill}
\small
\setlength{\tabcolsep}{3pt}
\begin{tabular}{@{}p{0.26\textwidth}p{0.31\textwidth}p{0.31\textwidth}@{}}
\toprule
Configuration & ODE & DMD2 \\
\midrule
Global Batch Size & 32 & 32 \\
Real-score CFG Weight & -- & 3.0 \\
Optimizer &
AdamW &
AdamW \\
Learning Rate ($G_\theta$) & $2.0{\times}10^{-6}$ & $2.0{\times}10^{-6}$ \\
Learning Rate ($p_{\mathrm{fake}}$) & -- & $4.0{\times}10^{-7}$ \\
Learning Rate Schedule ($G_\theta$ and $p_{\mathrm{fake}}$) & Cosine & Cosine \\
Learning Rate Warmup Step & -- & -- \\
Gradient Clipping ($G_\theta$ and $p_{\mathrm{fake}}$) & 10.0 & 10.0 \\
LoRA Rank ($G_\theta$) & 256 & 256 \\
LoRA Rank ($p_{\mathrm{fake}}$) & -- & 256 \\
LoRA Alpha ($G_\theta$) & 256 & 256 \\
LoRA Alpha ($p_{\mathrm{fake}}$) & -- & 256 \\
TTUR & -- & 5 \\
GAN Head Layers & -- & 5, 15, 25, 35, 39 \\
GAN Head Dim & -- & 768 \\
GAN Start Step & -- & 1000 \\
EMA Decay & 0.99 & 0.99 \\
EMA Start Step & 250 & 750 \\
Training Steps & 3759 & 2250 \\
Numerical Precision & BFloat16 & BFloat16 \\
GPU Usage & 32 NVIDIA H20 & 32 NVIDIA H20 \\
\bottomrule
\end{tabular}
\end{table*}

\subsection{System Efficiency Optimization}
\label{sec:Efficiency_Optimization}

To further accelerate both training and inference, we replace several memory-bound PyTorch operators with custom kernels and distributed execution strategies. These optimizations target the dominant cost centers of the AR diffusion transformer, including normalization, rotary position embedding, attention, VAE decoding, and long-sequence parallelism. Unless otherwise specified, the optimized kernels support both forward and backward propagation, allowing the same implementation to be used during large-scale training.
%and rollout-time generation.

\textbf{SageAttention.}
Attention dominates the runtime of long-window video diffusion transformers. We integrate SageAttention as a drop-in backend for compatible training and inference settings. By improving memory access patterns and using optimized low-precision execution, SageAttention reduces attention overhead and enables larger history memories and denser control tokens under the same GPU budget.

\textbf{TinyVAE decoding.}
Full VAE decoding is costly during rollout inspection. We therefore use a lightweight TinyVAE decoder based on TAESD~\citep{taesd} for preview decoding, while retaining the full VAE decoder for final evaluation and paper-quality visualization. This allows fast qualitative inspection during development without changing the final evaluation protocol.

\textbf{Ulysses sequence parallelism.}
To support long-horizon rollouts, we adopt Ulysses sequence parallelism and shard the token dimension across multiple GPUs. This reduces per-device activation storage approximately in proportion to the number of GPUs, allowing larger temporal windows and richer control conditions.

\textbf{Inference acceleration benchmark.}
We evaluate inference efficiency on an H20 96G GPU using $1920{\times}480$ videos with 99 frames. Attention-kernel optimization alone provides $1.25{\times}$--$1.31{\times}$ acceleration, while combining SageAttention with six-step DMD2 and Ulysses sequence parallelism reaches up to $35.93{\times}$ speedup.

% As shown in Figure~\ref{fig:efficiency_acceleration}, a

\textbf{Flash normalization.}
We fuse LayerNorm and RMSNorm in Triton by combining statistic computation, normalization, and affine transformation into a single kernel. Instead of materializing full normalized activations for backward propagation, the kernel caches only row-wise statistics, reducing intermediate memory usage. Internal reductions are computed in FP32, while inputs and outputs remain in the original training precision.

\textbf{Flash rotary position embedding.}
We implement a fused Triton kernel for Rotary Position Embedding (RoPE), avoiding repeated reshape, chunk, and indexing operations. The backward pass reuses the same kernel with the inverse rotation. By retaining only the pre-computed sine and cosine tables, the fused RoPE kernel reduces activation memory footprint and improves throughput for long-sequence rollout generation.

\subsection{Final Evaluation of GigaWorld-1}
\label{sec:Experiments}

We evaluate the final \textit{GigaWorld-1} from multiple perspectives to validate our design choices. Overall, the experiments confirm the effectiveness of the core components of \textit{GigaWorld-1}: the robot-oriented architecture improves benchmark performance, mixed training data enhances generalization, direct control leads to more accurate trajectories, and the final model produces more stable long-horizon rollouts than general-purpose video generation baselines.

\begin{figure*}[t]
\centering
\includegraphics[width=0.95\textwidth]{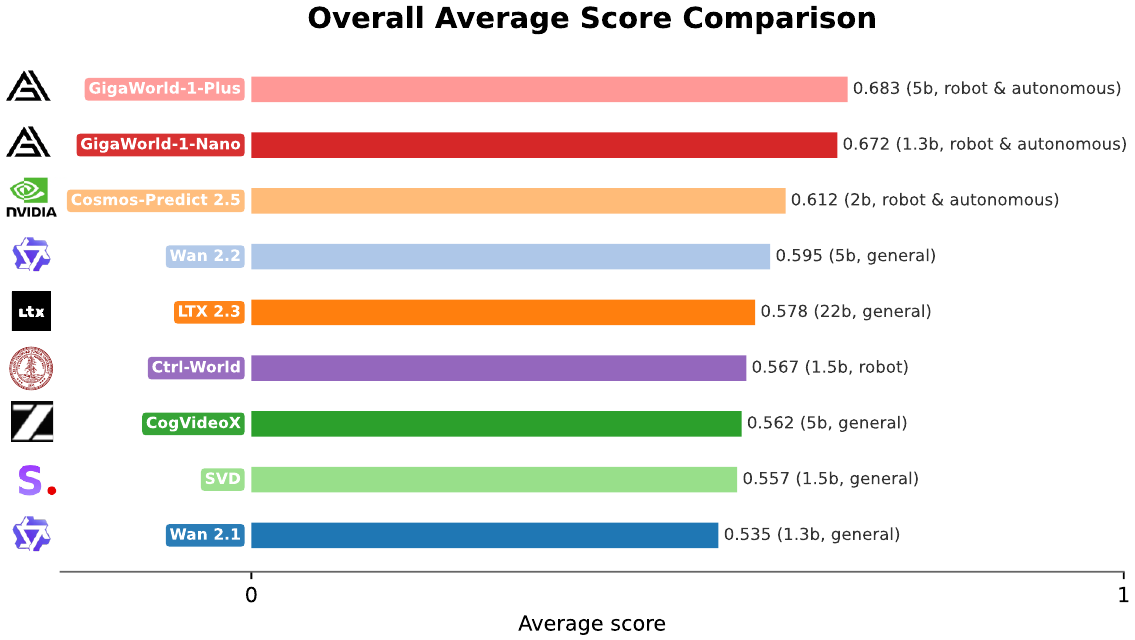}
\caption{\textbf{Model architecture comparison.} Left: mean score across six evaluation metrics. Right: radar plot of individual metric scores. \textit{GigaWorld-1-Nano} achieves the second-best overall score and performs particularly well in JEPA Similarity, Semantic Alignment, Subject Consistency, and Trajectory Accuracy.}
\label{fig:model_arch_compare}
\end{figure*}

\refstepcounter{paragraph}\paragraph*{\thesubsection.\arabic{paragraph}\quad Comparison with Open-Source Baselines}
We first compare \textit{GigaWorld-1} with representative general video generation and world-model baselines under a controlled setting: all backbones are post-trained on the same curated corpus described in Sec.~\ref{sec:data} and evaluated under the same rollout protocol. Following the metric taxonomy in Sec.~\ref{sec:wmbench_metrics} and the empirical findings in Sec.~\ref{sec:experiment}, we retain six core evaluator-relevant metric(Aesthetic Quality, Image Quality, JEPA Similarity, Semantic Alignment, Subject Consistency, and Trajectory Accuracy) and report their normalized average as the main summary score. We exclude appearance-stability metrics such as Background and Photometric Consistency, which can mislead evaluator ranking, and analyze long-horizon generation metrics separately in the rollout section. As shown in Figure~\ref{fig:model_arch_compare} and Table~\ref{tab:model_arch_comparison}, \textit{GigaWorld-1-Plus} achieves the best overall result with an average score of 0.6834, followed by \textit{GigaWorld-1-Nano} at 0.6717. Among the open-source baselines, Cosmos-Predict2.5 is the strongest (0.6123), while Wan 2.2 5B is the strongest general-purpose model (0.5948). Compared with these two strongest baselines, \textit{GigaWorld-1-Plus} improves the average score by 11.6\% over Cosmos-Predict2.5 and by 14.9\% over Wan 2.2 5B. More importantly, the gains are concentrated on the evaluator-critical metrics identified in Question~I: \textit{GigaWorld-1-Plus} achieves the best JEPA Similarity (0.9337), Semantic Alignment (0.8926), and Trajectory Accuracy (0.3561), while matching the best Subject Consistency score (0.8883).

\begin{table*}[t]
\centering
\caption{\textbf{Model architecture comparison.} Robot-oriented models generally outperform generic video backbones on embodied rollout evaluation. Cell background colors indicate ranking: \cellcolor{rankbest}green~= 1st, \cellcolor{rankmid}yellow~= 2nd, \cellcolor{rankworst}red~= last. All metrics are higher-is-better ($\uparrow$).}
\label{tab:model_arch_comparison}
\scriptsize
\setlength{\tabcolsep}{3pt}
\resizebox{\textwidth}{!}{
\begin{tabular}{llcrrrrrrr}
\toprule
Model & Size & Type & Aesthetic$\uparrow$ & Image$\uparrow$ & JEPA$\uparrow$ & Semantic$\uparrow$ & Subject$\uparrow$ & Trajectory$\uparrow$ & AVG$\uparrow$ \\
\midrule
SVD & 1.5B & General 
& \cellcolor{rankworst}0.2861 
& \cellcolor{rankworst}0.6497 
& 0.6454 
& \cellcolor{rankworst}0.8411 
& 0.8267 
& \cellcolor{rankworst}0.0926 
& 0.5569 \\

Wan 2.1 1.3B I2V & 1.3B & General 
& 0.3422 
& 0.6856 
& 0.6002 
& 0.8705 
& \cellcolor{rankworst}0.5568 
& 0.1576 
& \cellcolor{rankworst}0.5355 \\

LTX 2.3 & 22B & General 
& \cellcolor{rankbest}\textbf{0.3900} 
& 0.6967 
& \cellcolor{rankworst}0.5380 
& 0.8678 
& 0.8248 
& 0.1479 
& 0.5775 \\

CogVideoX & 5B & General 
& 0.3303 
& 0.6775 
& 0.6437 
& 0.8633 
& 0.6963 
& 0.1609 
& 0.5620 \\

Wan 2.2 5B TI2V & 5B & General 
& \cellcolor{rankmid}0.3538 
& \cellcolor{rankmid}0.6980 
& 0.5853 
& 0.8789 
& \cellcolor{rankbest}\textbf{0.8883} 
& 0.1643 
& 0.5948 \\

Cosmos-Predict2.5 & 2B & Robot/Auto 
& 0.3491 
& \cellcolor{rankbest}\textbf{0.7184} 
& 0.6781 
& 0.8764 
& \cellcolor{rankmid}0.8747 
& 0.1770 
& 0.6123 \\
\midrule

\textit{GigaWorld-1-Nano} & 1.3B & Robot/Auto 
& \cellcolor{rankmid}0.3538 
& 0.6802 
& \cellcolor{rankmid}0.8911 
& \cellcolor{rankmid}0.8920 
& 0.8600 
& \cellcolor{rankmid}0.3528 
& \cellcolor{rankmid}0.6717 \\

\textit{GigaWorld-1-Plus} & 5B & Robot/Auto 
& 0.3534 
& 0.6765 
& \cellcolor{rankbest}\textbf{0.9337} 
& \cellcolor{rankbest}\textbf{0.8926} 
& \cellcolor{rankbest}\textbf{0.8883} 
& \cellcolor{rankbest}\textbf{0.3561} 
& \cellcolor{rankbest}\textbf{0.6834} \\
\bottomrule
\end{tabular}
}
\end{table*}

\begin{figure*}[t]
\centering
\includegraphics[width=0.95\textwidth]{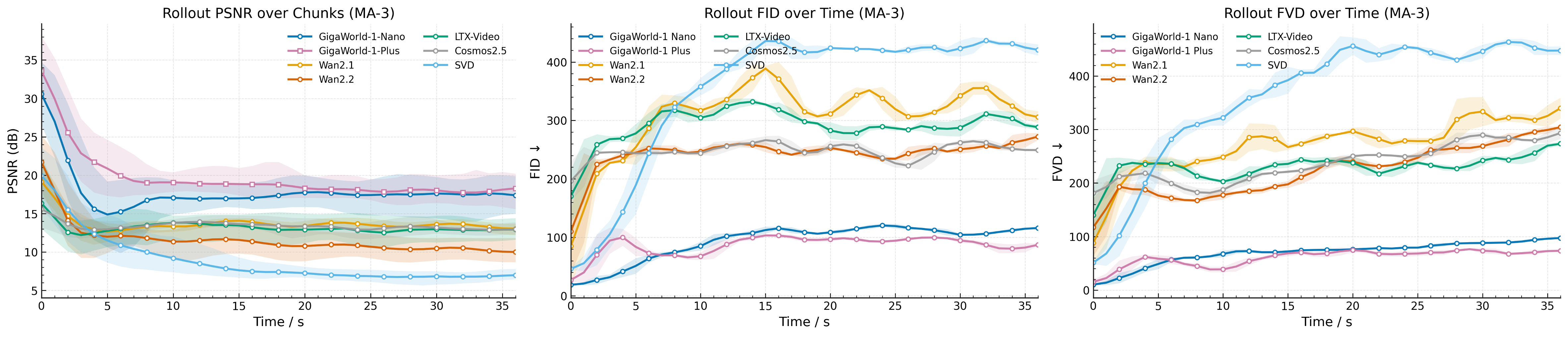}
\caption{\textbf{Long-horizon rollout dynamics.} PSNR ($\uparrow$) and FID ($\downarrow$) are measured over successive 10-frame rollout chunks, showing how reconstruction fidelity and perceptual quality evolve with rollout length.}
\label{fig:rollout_psnr_fid}
\end{figure*}

\begin{figure*}[t]
\centering
\includegraphics[width=0.95\textwidth]{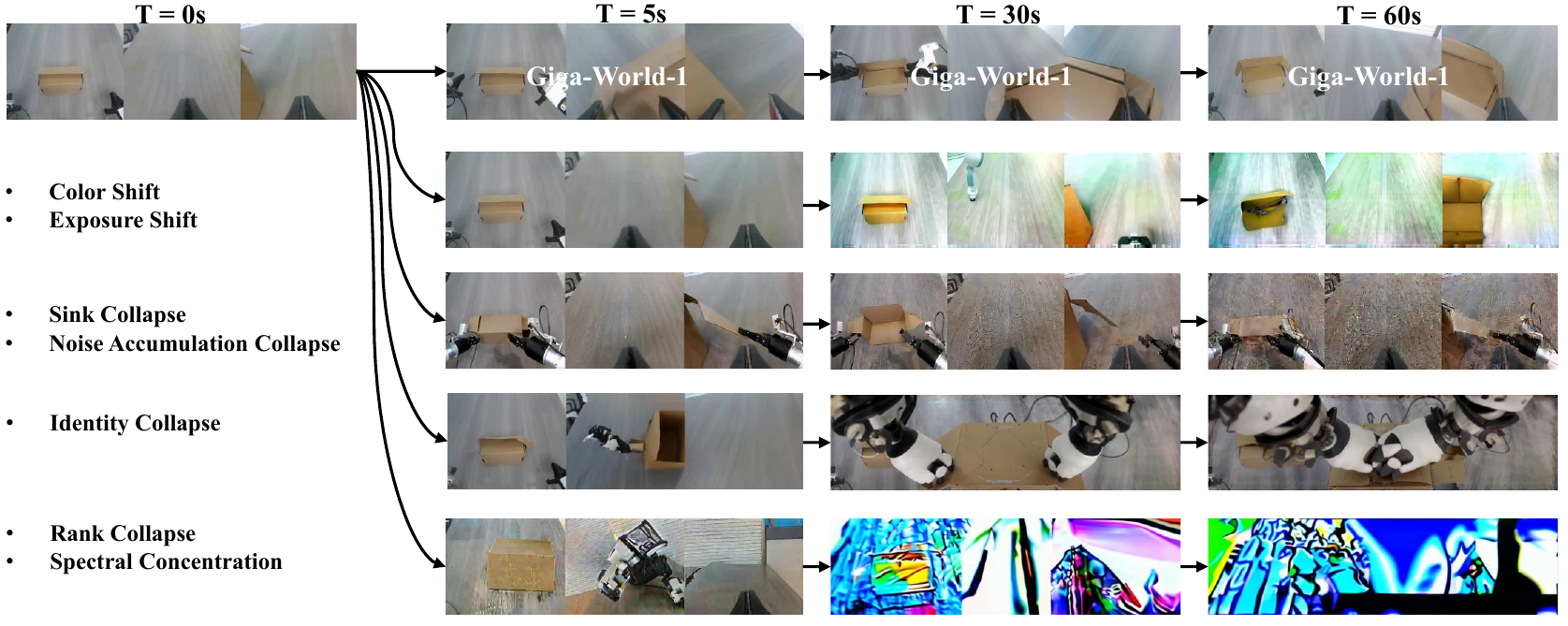}

\caption{\textbf{Long-horizon model comparison.} \textit{GigaWorld-1} is compared with general video generation and world-model baselines under the same rollout evaluation protocol.}
\label{fig:rollout_model_compare}
\end{figure*}

\refstepcounter{paragraph}\paragraph*{\thesubsection.\arabic{paragraph}\quad Long-Horizon Rollout Quality}
We evaluate long-horizon video rollout quality chunk by chunk, using 10-frame chunks corresponding to approximately 1 second. PSNR is computed across multi-view settings, while FID and FVD are computed on the head view. As shown in Figure~\ref{fig:rollout_psnr_fid} and Table~\ref{tab:rollout_quality_exp}, \textit{GigaWorld-1} consistently achieves the best PSNR, FID, and FVD over 40 seconds of autoregressive generation. Qualitatively, generic baselines suffer from viewpoint drift, object-identity collapse, and accumulated texture artifacts, as shown in Figure~\ref{fig:rollout_model_compare} (more videos are shown in the project page). In contrast, \textit{GigaWorld-1} avoids these failure modes by using hierarchical history memory and unified control injection (Sec.~\ref{sec:model_structure}), which explicitly constrain robot motion and preserve long-term scene identity.

\refstepcounter{paragraph}\paragraph*{\thesubsection.\arabic{paragraph}\quad Impact of Memory and Prompt Interpolation}

\begin{figure*}[t]
\centering
\includegraphics[width=0.95\textwidth]{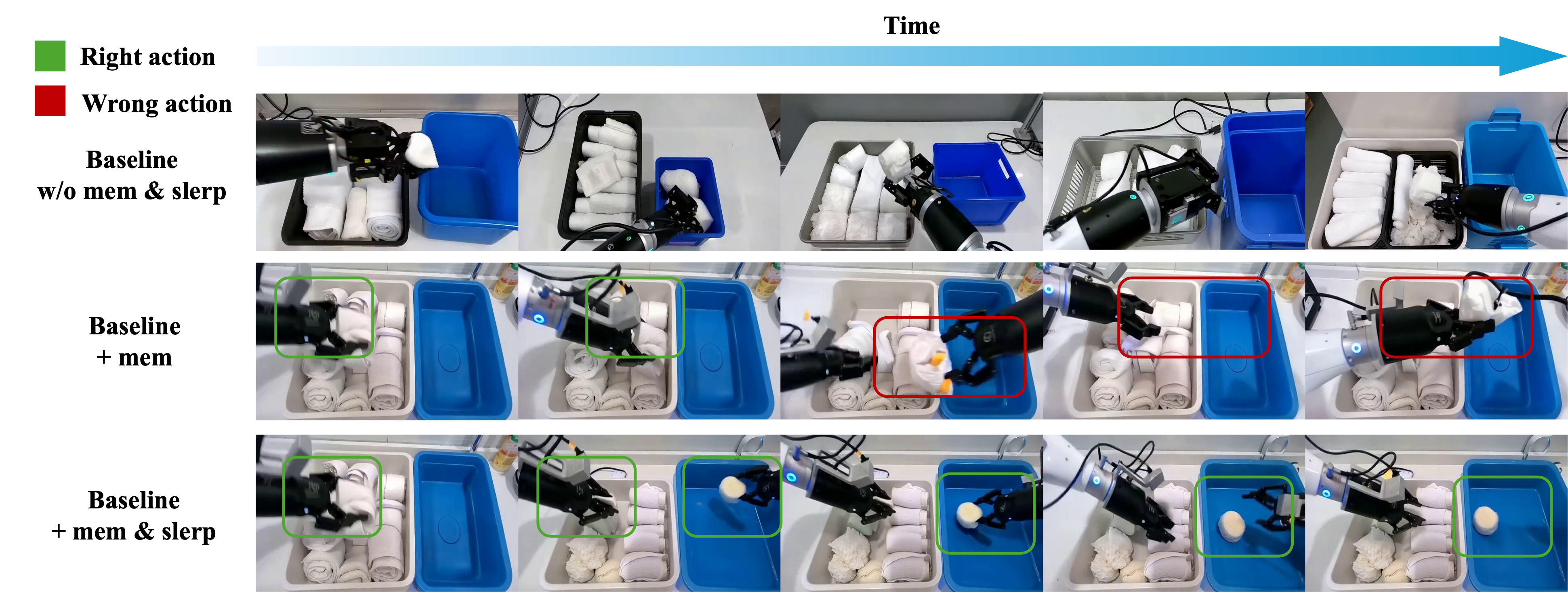}
\caption{\textbf{Effect of memory and prompt interpolation} (left arm moves to a predefined observation pose, while the right arm places a towel into the blue box). Without memory, the rollout suffers from background jumps and inconsistent scene layout. Adding history memory stabilizes the workspace, but a fixed or abruptly switched prompt can make the model over-conditioned on the previous subtask during stage transitions. Combining memory with prompt interpolation preserves background consistency while enabling smoother semantic transition from the left-arm observation pose to the right-arm towel-placement action.}
\label{fig:memory_prompt_interpolation}
\end{figure*}

We conduct a qualitative ablation to isolate the effect of historical memory and prompt interpolation in a multi-stage dual-arm manipulation task (Figure~\ref{fig:memory_prompt_interpolation}). Without memory, the rollout suffers from background jumps and inconsistent scene layouts. While adding memory stabilizes the workspace, abrupt prompt changes can cause the model to over-condition on previous subtasks. Combining memory with Spherical Linear Interpolation (SLERP) resolves this trade-off, enabling smoother semantic transitions from one task phase to the next.

\begin{figure*}[t]
\centering
\includegraphics[width=0.95\textwidth]{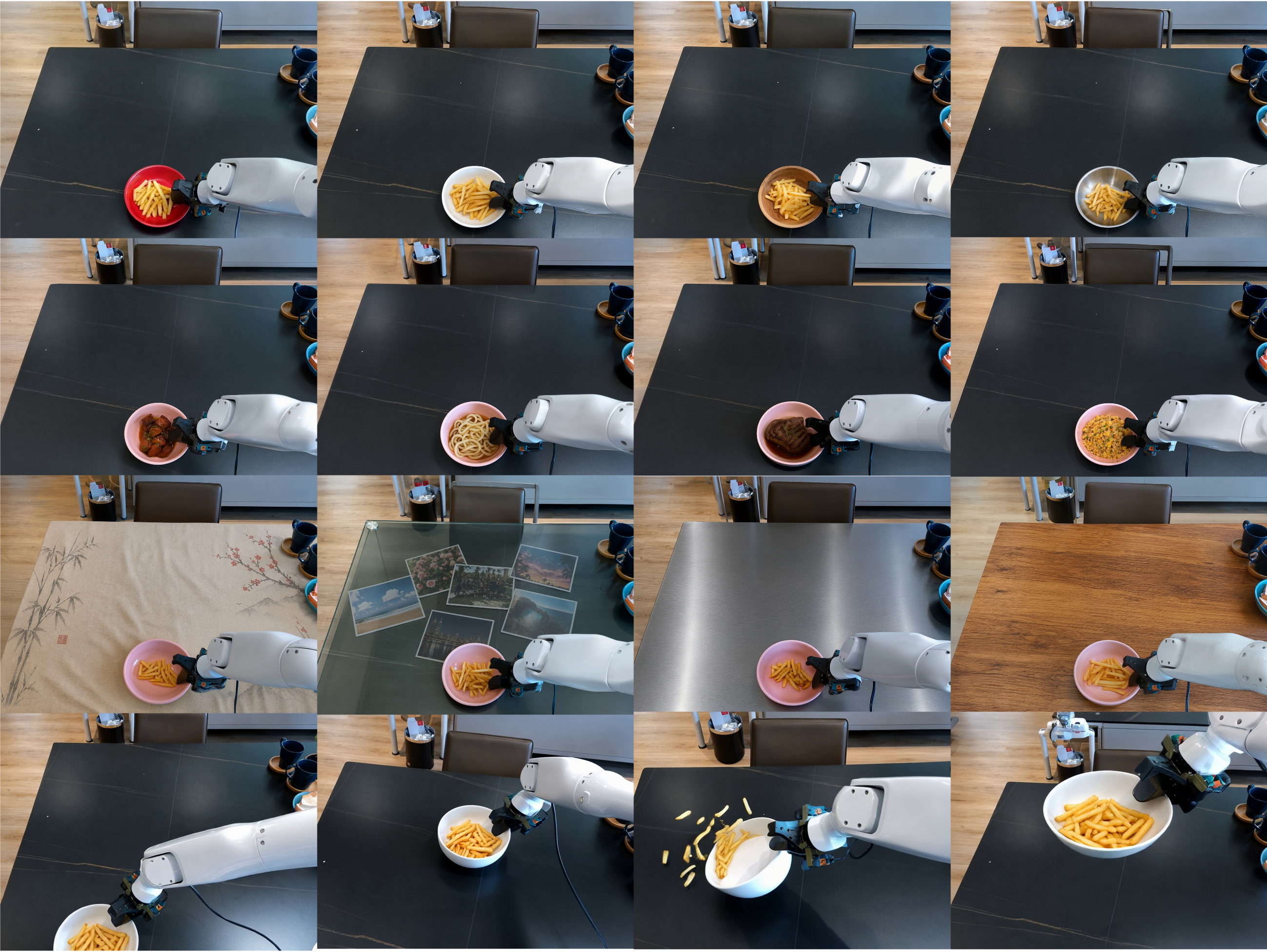}
\caption{\textbf{OOD generalization cases.} From top to bottom, the rows evaluate generalization to object color and container appearance, object content changes, background and table-surface changes, and action-outcome variation covering both successful and failed executions.}
\label{fig:ood_generalization}
\end{figure*}

\refstepcounter{paragraph}\paragraph*{\thesubsection.\arabic{paragraph}\quad OOD Generalization}
We evaluate out-of-distribution (OOD) generalization by testing whether the world model preserves physical behavior under shifts in object appearance, object category, scene background, and policy outcome. As shown in Figure~\ref{fig:ood_generalization}, \textit{GigaWorld-1} successfully handles changes in container color, object contents (e.g., different food types), and table textures without geometry collapse. Crucially, the fourth row demonstrates action-outcome generalization: the model accurately simulates both successful placements and failures (e.g., spilling), which is an essential property for a reliable policy evaluator.

\refstepcounter{paragraph}\paragraph*{\thesubsection.\arabic{paragraph}\quad Closed-Loop Policy Consistency}
Replay fidelity alone does not imply a world model can evaluate policies. In closed-loop use, small errors in visual state or contact can compound and alter the task outcome. We evaluate whether \textit{GigaWorld-1} preserves the same success/failure conclusion as physical-robot execution using the hybrid VLM-evaluator on the WMBench closed-loop tasks (Table~\ref{tab:closed_loop_tasks}).

\begin{figure*}[t]
\centering
\includegraphics[width=1\textwidth]{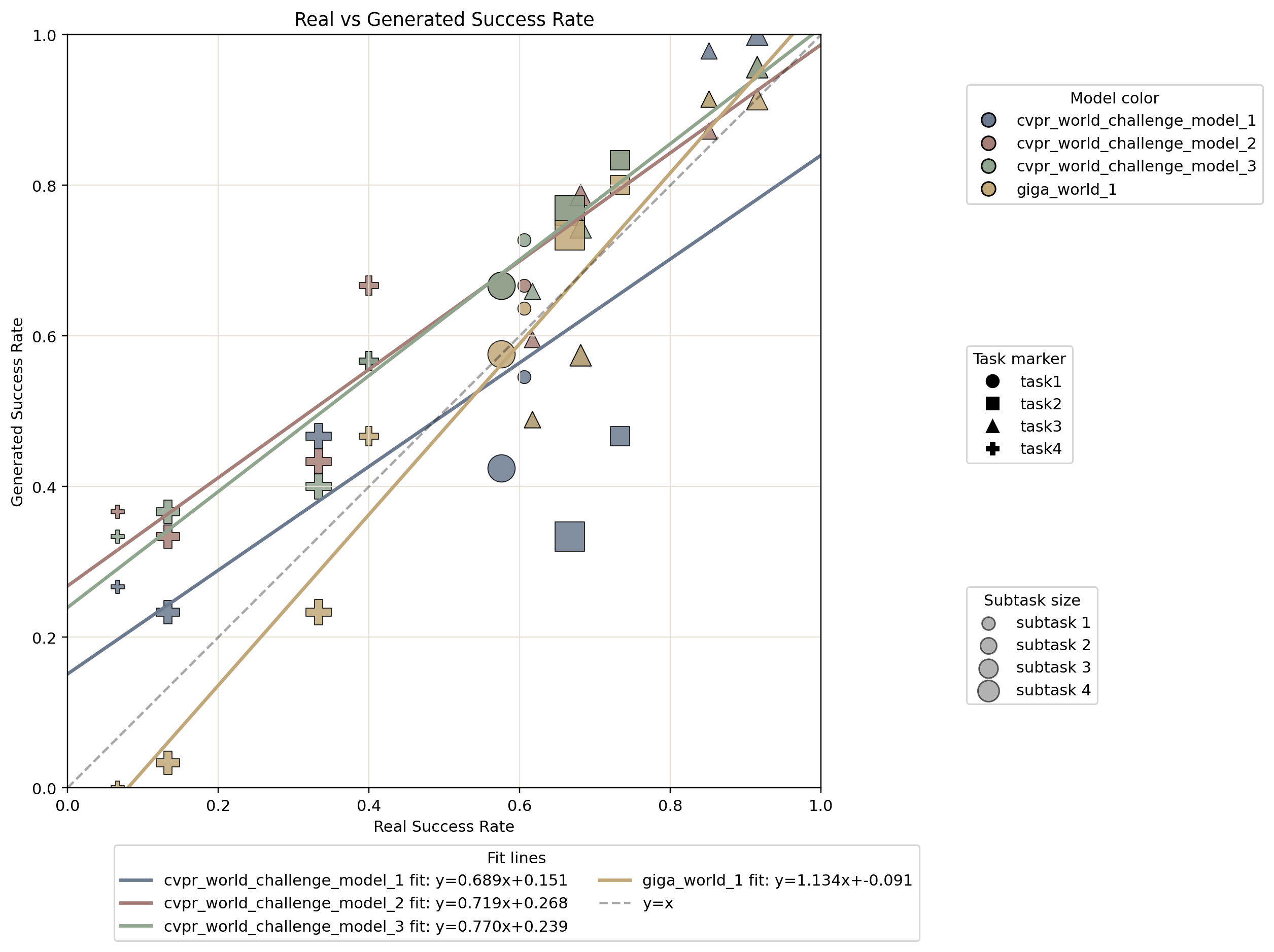}
\caption{\textbf{Task-level success-rate alignment.} Real-robot success rates are compared with generated success rates under closed-loop policy rollout. The gray dashed line indicates perfect agreement. \textit{GigaWorld-1} has a fitted line closer to the diagonal than the challenge baselines, indicating better calibration of task difficulty across the evaluated tasks and subtasks.}
\label{fig:closed_loop_success_fit}
\end{figure*}

As shown in Figure~\ref{fig:closed_loop_success_fit}, task-level success-rate alignment provides a direct measure of evaluator calibration. The challenge baselines often overestimate policy success, reflecting a common bias where generated rollouts look visually plausible but fail to penalize poor actions. In contrast, \textit{GigaWorld-1} follows the real-world diagonal much more closely, indicating better preservation of relative task difficulty.

\begin{table*}[t]
\centering
\caption{\textbf{Closed-loop policy consistency tasks.} Each task is decomposed into subtask-level outcome checks so that agreement can be measured not only at the task level, but also at contact- and manipulation-sensitive intermediate stages.}
\label{tab:closed_loop_tasks}
\resizebox{0.95\textwidth}{!}{
\begin{tabular}{lll}
\toprule
Task & Task Type & Subtasks \\
\midrule
task1 & Put banana into basket & Grasp banana; Place banana into basket \\
task2 & Put green bowl into pink plate & Grasp green bowl; Place green bowl into pink plate \\
task3 & Fold paper boxes & Grasp carton flaps; Gripper in position; Press down carton lids; Reset \\
task4 & Pour the fries into the box & Move the takeout box over the plate; Open the takeout box; Pour out the fries; Press down on the takeout box \\
\bottomrule
\end{tabular}
}
\end{table*}

\begin{figure*}[t]
\centering
\includegraphics[width=1\textwidth]{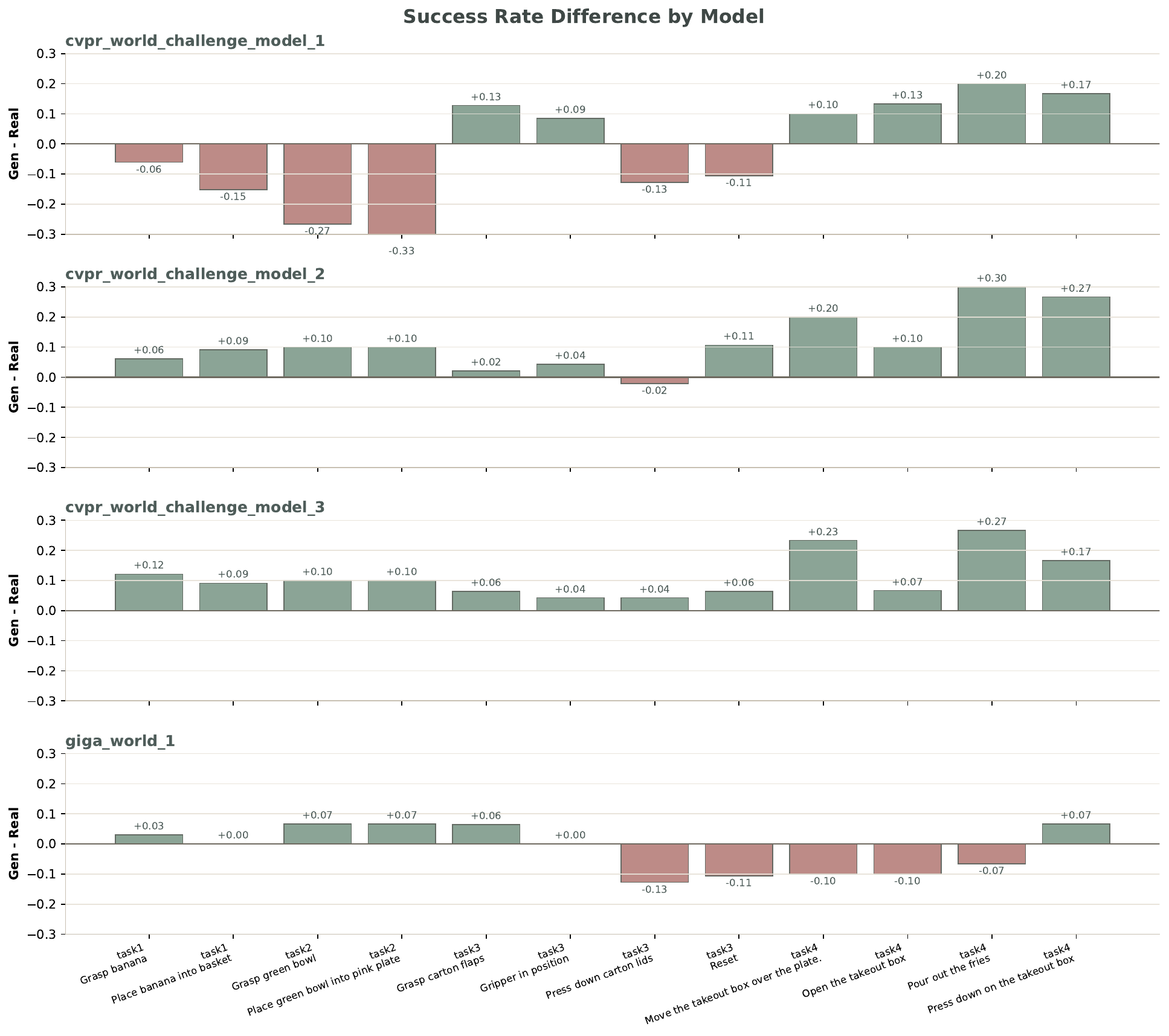}
\caption{\textbf{Success-rate bias across world models and subtasks.} Bars show $\mathrm{Gen} - \mathrm{Real}$ success-rate differences: green indicates overestimation of real-world success, red indicates underestimation, and values near zero indicate closer agreement.}
\label{fig:closed_loop_confusion}
\end{figure*}

Success-rate bias analysis (Figure~\ref{fig:closed_loop_confusion}) further compares generated and real robot outcomes across manipulation subtasks. \textit{GigaWorld-1} shows smaller overall $\mathrm{Gen} - \mathrm{Real}$ deviations than \textit{cvpr\_world\_challenge\_model\_1}, suggesting closer calibration to real-world success rates. In contrast, \textit{cvpr\_world\_challenge\_model\_2} and \textit{cvpr\_world\_challenge\_model\_3} tend to over-confidently predict success, whereas \textit{GigaWorld-1} provides a more balanced estimate across subtasks.

 Therefore, closed-loop policy consistency is stricter than replay quality. A useful world model must be calibrated not only to visual realism, but also to the success and failure distribution induced by a policy interacting with the generated state. The results show that \textit{GigaWorld-1} better preserves closed-loop policy conclusions than generic video-generation baselines, while also highlighting an important direction for future work: reducing the optimistic bias of video generation models on contact-sensitive failure.

%% file: sections/discussion.tex
\section{Discussion and Conclusion}

Our study suggests that the central difficulty of learned robot evaluation is not video generation alone, but preserving \emph{policy-relevant causality} under iterative rollout. This insight explains why seemingly strong video models can still be poor evaluators: they may generate plausible local motion while drifting away from the action-conditioned evolution required for reliable policy comparison. More broadly, \textit{GigaWorld-1} explores how to build a world model that can serve as a reliable policy evaluator under a cost-controllable recipe. In this setting, evaluator design cannot be reduced to a single axis such as model size or robot-data volume alone. What matters is the ability to preserve broad world knowledge, remain controllable under action input, and sustain long-horizon consistency while keeping the overall evaluation pipeline practical enough for repeated use.

The work also has practical implications for benchmark design and deployment. If future robot foundation models are to iterate as quickly as digital foundation models, the community needs reliable, reusable evaluator infrastructure. WMBench is intended as a step in this direction: not merely another benchmark with a fixed leaderboard, but a tool for studying why evaluator conclusions succeed or fail. We do not encourage optimizing narrowly for leaderboard gains on WMBench. Instead, the broader goal is to use large-scale experimentation to identify which modeling choices and which evaluation signals are truly reliable for world models as policy evaluators. Beyond benchmark analysis, we have already integrated \textit{GigaWorld-1} into our internal policy-evaluation pipeline and conducted extensive testing at scale. This experience suggests that world-model-based evaluation can already provide meaningful support for policy iteration in practice, and that continued accumulation of rollout data should further strengthen \textit{GigaWorld-1} as both a simulator and an evaluator. The public challenge component further demonstrates that community-driven model exploration can enrich this research question by broadening the diversity of evaluator candidates.

At the same time, our conclusions should be interpreted with several limitations in mind. First, although WMBench covers eight task families, it does not yet exhaust the full space of mobile manipulation, dexterous in-hand manipulation, or safety-critical autonomy. Second, our study focuses primarily on video-centric world models; other structured state-space or hybrid 3D approaches may exhibit different trade-offs. Third, while VLM-assisted success labeling substantially reduces annotation cost, final evaluator assessment still benefits from human verification in uncertain cases. Looking ahead, several directions appear especially promising for world models as simulators and policy evaluators. We also plan to continue expanding the benchmark with more data, especially more rollout data, so that evaluator analysis can be grounded in broader and harder closed-loop settings. In parallel, more accurate metrics for measuring outcome faithfulness, action-conditioned consistency, and evaluator reliability remain important for making world-model-based policy evaluation more trustworthy. Recent community efforts on omni world models, such as Cosmos-3~\citep{cosmos3}, suggest that absorbing richer multimodal training data may improve both controllability and physical grounding. In parallel, scaling model parameters remains an important direction, not as a substitute for evaluator-oriented design, but as a complementary path for improving capacity, world knowledge retention, and long-horizon robustness. Altogether, we hope these efforts will accelerate the adoption of world models as policy evaluators in both research and industrial practice, while extending these results toward richer reward modeling, counterfactual evaluation, and certified uncertainty.